\documentclass[10pt,journal,compsoc]{IEEEtran}
\ifCLASSOPTIONcompsoc
    \usepackage[noadjust]{cite}
\else
  \usepackage[noadjust]{cite}
\fi
\ifCLASSINFOpdf
  
\else
  
\fi

\usepackage{amsmath}
\usepackage{url}
\usepackage{xcolor}
\usepackage{times}
\usepackage{epsfig}
\usepackage{graphicx}
\usepackage{caption}
\usepackage{float}
\usepackage{amssymb} 
\usepackage{footmisc}
\usepackage{lineno}
\usepackage{color}
\usepackage{subfigure}
\usepackage{multirow}
\usepackage{dsfont}
\usepackage{mathtools}
\usepackage{setspace}
\usepackage{csquotes}
\usepackage{breqn}
\usepackage{lettrine}
\usepackage{upquote}
\usepackage{adjustbox}
\usepackage{wrapfig}
\usepackage{lipsum}
\usepackage{mathtools}
\usepackage{comment}
\usepackage{bm}
\usepackage{array}
\usepackage{tabu}
\usepackage{booktabs}
\usepackage{enumerate}
\DeclareCaptionFormat{myformat}{\fontsize{8}{10}\selectfont#1#2#3}
\DeclareMathOperator*{\argmin}{argmin}
\DeclareMathAlphabet\mathbfcal{OMS}{cmsy}{b}{n}
\DeclareMathOperator*{\argmax}{argmax}
\DeclareMathAlphabet{\pazocal}{OMS}{zplm}{m}{n}
\DeclareMathAlphabet{\mathpzc}{OT1}{pzc}{m}{it}

\usepackage[ruled,vlined]{algorithm2e}
\usepackage[colorlinks=true,linkcolor=blue]{hyperref}
\def\arrvline{\hfil\kern\arraycolsep\vline\kern-\arraycolsep\hfilneg}

\ifCLASSINFOpdf
\else
\fi

\begin{document}
\bstctlcite{IEEEexample:BSTcontrol}
\title{Visual Object Tracking with Discriminative Filters and Siamese Networks: A Survey and Outlook}

\author{Sajid Javed,
Martin Danelljan,
Fahad Shahbaz Khan, ~\IEEEmembership{IEEE Senior Member},
Muhammad Haris Khan,
Michael Felsberg, ~\IEEEmembership{IEEE Senior Member},
~and Jiri Matas~\IEEEmembership{IEEE Senior Member}
\IEEEcompsocitemizethanks{\IEEEcompsocthanksitem S.\ Javed is with the EECS department, Khalifa University of Science and Technology, P.O Box : 127788, Abu Dhabi, UAE. (email: sajid.javed@ku.ac.ae).
\IEEEcompsocthanksitem M. Danelljan is with the Computer Vision Lab, Dept.\ of Information Technology and Electrical Engineering, ETH Z\"urich, Switzerland.
\IEEEcompsocthanksitem F.\ S.\ Khan and M.\ H.\ Khan are with computer vision department, MBZUAI, Abu Dhabi, UAE.
\IEEEcompsocthanksitem M.\ Felsberg and F.\ S.\ Khan are with computer vision laboratory, Linköping University, Sweden.
\IEEEcompsocthanksitem J.\ Matas is with Center for Machine Perception, Czech Technical University, Prague.}}

\markboth{Journal of \LaTeX\ Class Files,~Vol.~14, No.~8, August~2015}
{Javed \etal: VOT}

\IEEEtitleabstractindextext{

\begin{abstract}
Accurate and robust visual object tracking is one of the most challenging and fundamental computer vision problems. It entails estimating the trajectory of the target in an image sequence, given only its initial location, and segmentation, or its rough approximation in the form of a bounding box. 
Discriminative Correlation Filters (DCFs) and deep Siamese Networks (SNs) have emerged as dominating tracking paradigms, which have led to significant progress. 
Following the rapid evolution of visual object tracking in the last decade, this survey presents a systematic and thorough review of more than 90 DCFs and Siamese trackers, based on results in nine tracking benchmarks. 
First, we present the background theory of both the DCF and Siamese tracking core formulations. 
Then, we distinguish and comprehensively review the shared as well as specific open research challenges in both these tracking paradigms. Furthermore, we thoroughly analyze the performance of DCF and Siamese trackers on nine benchmarks, covering different experimental aspects of visual tracking: datasets, evaluation metrics, performance, and speed comparisons. We finish the survey by presenting recommendations and suggestions for distinguished open challenges based on our analysis.
\end{abstract}

}

\maketitle

\IEEEpeerreviewmaketitle
\section{Introduction}
\noindent \lettrine[lraise=0.1, nindent=0em, slope=-.5em]{V}{isual} Object Tracking (VOT) is one of the fundamental open problems in computer vision. 
The task is to estimate the trajectory and state of a target in an image sequence.
VOT has a wide range of applications, including autonomous driving, robotics, intelligent video surveillance, sports analytics and medical imaging, where it typically plays an important role within large intelligent systems. 
Given the initial state of any arbitrary target object, the main challenge in VOT is to learn an appearance model to be used when searching for the target object in subsequent frames. 
In recent years, VOT has received considerable attention, much thanks to the introduction of a variety of tracking benchmarks such as, TackingNet \cite{muller2018trackingnet}, VOT2018 \cite{kristan2018sixth}, and GOT-10K \cite{huang2019got}. 
Despite the recent progress, VOT is 
still an open research problem and is perhaps more active than ever \cite{fiaz2019handcrafted, li2013survey}.

\begin{figure}[t!]
\centering
\includegraphics[width=\linewidth]{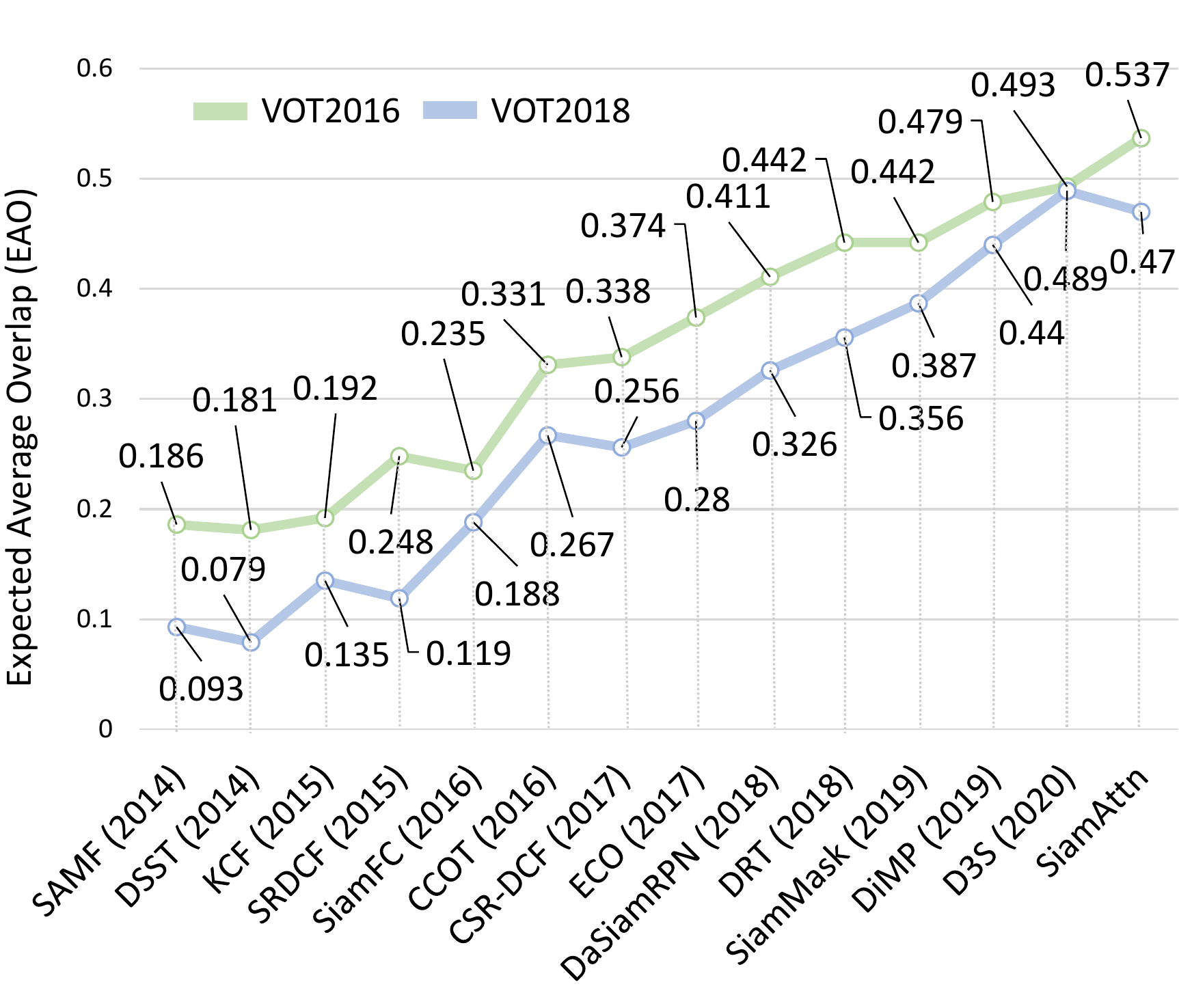}
\vspace{-1.7\baselineskip}
\caption{Performance improvements from 2014 to 2020 on the popular VOT benchmark series. Tracking quality is measured in terms of Expected Average Overlap (EAO). We show DCF and Siamese trackers that significantly advanced the SOTA. 
The representative trackers are: SAMF \cite{SAMF14}, DSST \cite{DSST14}, KCF \cite{KCF15}, SRDCF \cite{SRDCF15}, SiamFC \cite{SIAMFC16},  CCOT \cite{CCOT16}, CSR-DCF \cite{CSRDCF17}, ECO \cite{ECO17}, DaSiamRPN \cite{DASIAMRPN18}, DRT \cite{DRT18}, SiamMask \cite{SIAMMASK19}, DiMP \cite{DIMP19}, D3S \cite{D3S20} and SiamAttn \cite{SIAMATTN20}.}
\label{fig_VOT}
\vspace{-1.5\baselineskip}
\end{figure}

The core challenge in generic object tracking\footnote{Generic object tracking refers to the problem, where the tracked object is not known a priori and  it is not constrained to be from a specific class.} is to learn an appearance model of an arbitrary target object online, given only its initial state. 
Several real-world factors, complicates learning an accurate appearance model.
For instance, the target object may undergo a  partial or full occlusion, scale variation, and deformation. Moreover, there are environmental factors, including illumination changes and motion blur, which influence the appearance of the target. Another factor is that scenes often include objects or background structures having similar appearance that can easily be confused with the target itself \cite{wu2015object}. 
To address these challenges, a plethora of trackers have been proposed in the literature that have contributed to advance the State-Of-The-Art (SOTA) in tracking. 

In the past decade, Discriminative Correlation Filters (DCFs) and deep Siamese Networks (SNs) have been the two most prominent paradigms for VOT \cite{fiaz2019handcrafted , SIAMFC16, SINT16, GOTURN16}.
In DCF-based tracking, a correlation filter is trained online on the region of interest by minimizing a least squares loss. 
The target is then detected in consecutive frames by convolving the trained filter via the fast Fourier transform (FFT) \cite{MOSSE10,KCF15}. 
In the deep Siamese tracking framework, an embedding space is learned offline by maximizing the distance between the target and background appearance, while minimizing the distance between the two patches from the target itself. 
The SNs consists of two identical sub-branches: one for the target template and the other for the target search region. 
The network takes both template and search regions as inputs and outputs the local similarity with the target for each location in the search region. 
The SNs were initially proposed for the signature verification task \cite{bromley1993signature} and later explored for visual tracking by Tao \textit{et al.} \cite{SINT16} and Bertinetto \textit{et al.} \cite{SIAMFC16}. 
With the design of DCFs \cite{CSK12,STC14, ACA14} and SNs \cite{SIAMFC16, GOTURN16, SINT16}, the tracking community has been largely focusing on these two paradigms in recent years as these two frameworks have significantly improved the tracking performance on several datasets such as the performance improvements on VOT datasets \cite{hadfield2016visual, kristan2018sixth} as shown in Fig. \ref{fig_VOT}. 

In this work, we present a systematic review of popular DCF and Siamese-based tracking paradigms.
Both paradigms share the same objective, that is to learn an accurate target appearance model that can effectively discriminate the target object from the background. 
While emerging from different underlying paradigms for addressing the aforementioned objective, the advent of deep learning has brought several important similarities and common challenges to these two paradigms. 
For instance, (i) \textit{Feature Representation}: Both paradigms exploit different features representations to estimate target translations and scale variations. 
Leveraging deep feature representations extracted from pre-trained networks is a recent trend shared by both paradigms. 
However, the choice of deep architecture and feature hierarchies is still an open problem in both tracking paradigms.
(ii) \textit{Target State Estimation}: The core formulation of both DCF and Siamese trackers only addresses how to estimate the translation of the target object. Hence, neither paradigm provides an explicit method for estimating the full target state, parametrized by e.g.\ a bounding box, which is crucial in most applications. 
(iii) \textit{Offline Training}: While initially only Siamese trackers benefited from end-to-end offline training, recent DCF trackers \cite{ATOM19, DIMP19,PRDIMP20} also leverage large-scale offline learning, integrating it with efficient and differentiable online learning modules for robust and accurate tracking.

While having common attributes, these two popular paradigms also have specific issues.
For instance, (i) \textit{Boundary Artifacts}: DCF-based trackers generally exploit the periodic assumption of training samples for learning an online classifier, which introduces undesirable boundary effects that severely degrade the quality of the target model. 
(ii) \textit{Optimization:} The loss function minimization also introduces challenges in DCF-based trackers, especially when target-specific constraints, such as spatial or temporal, etc., are regularized within the regression loss. 
(iii) \textit{Online Model Adaptability}: When the target appearance undergoes changes due to variation in lighting conditions or fast motion etc., the learned model is expected to cope with these variations.
The DCF trackers possess the ability to update the appearance model over time through the loss function. On the other hand, Siamese trackers do not inherent such a mechanism for online model update. Therefore, online adaptability is an important issue in Siamese trackers.
\noindent In the paper, we systematically review the popular DCF and Siamese tracking paradigms with the following contributions:

\begin{itemize}
\item We distinguish and thoroughly review the shared as well as specific open challenges in both tracking paradigms. 
\item We present an extensive background theory of both DCF and Siamese tracking core formulations. 
\item We present a comprehensive overview of more than 90 DCF and Siamese-based trackers in the literature, presenting a comparison on nine tracking benchmarks.
\item Based on our analysis, we present our recommendations and suggestions for specific open challenges. 
\end{itemize}

\noindent To the best of our knowledge, we are the first to present a comprehensive survey of the two most popular tracking paradigms (DCF and Siamese) in the last decade, describing their respective background theory, specifying their shared as well as specific open research issues along with providing a set of recommendations and future directions.  

The rest of this paper is organized as follows:
Section \ref{sec:literature} focuses on other surveys on tracking.
In Sections \ref{sec:maindcf} and \ref{sec:mainsiamese}, we 
describe the core DCF and Siamese tracking formulation, respectively. 
Then, we present a comprehensive overview of DCF and Siamese-based trackers, respectively, by distinguishing open key research issues. Afterwards, we provide an overview of the evolution of these two paradigms into segmentation-based tracking frameworks.
Experimental evaluations are presented in Sections \ref{sec:mainresults}.
We conclude our survey and provide further research directions in Section \ref{sec:maindiscussion}.

\section{Literature Review}
\label{sec:literature}
In the literature, numerous survey studies on VOT  have been published in the past two decades \cite{yilmaz2006object, yang2011recent, smeulders2013visual, li2018deep, fiaz2019handcrafted, wang2019comparison, li2013survey, Felsberg2021, zhang2013sparse}. 
In the first survey, Yilmaz \textit{et al.} presented systematic analysis of the whole tracking procedure \cite{yilmaz2006object}. 
The tracking methods were categorized into: based on the point or features correspondence, primitive geometric models, and contour methods. 
Smeulders \textit{et al.} presented an experimental survey of 19 different online trackers that emerged from 1999 to 2012 with multiple evaluation metrics on a newly proposed ALOV++ dataset \cite{smeulders2013visual}. 
Zhang \textit{et al.} presented sparse coding-based trackers survey in which the tracking methods were categorized into sparse coding and sparse representation-based trackers \cite{zhang2013sparse}. 
Recently, Li \textit{et al.} presented a survey of deep trackers and provided an experimental comparison \cite{li2018deep}. 
The trackers were classified and evaluated in terms of network structure, network function, and network training. 
Fiaz \textit{et al.} reviewed SOTA DCFs and non DCF-based trackers, providing comparative study based on the feature extraction methods \cite{fiaz2019handcrafted}.

The main differences between this survey and previous VOT surveys \cite{yilmaz2006object, li2018deep, fiaz2019handcrafted , zhang2013sparse} are as follows. 
Unlike previous VOT surveys, our work focuses solely on the two best-performing tracking paradigms, DCFs and SNs, in recent years. 
We present an extensive background theory of both DCF and Siamese tracking core formulations. 
We then provide an extensive overview of more than 90 DCF and Siamese trackers and the evolution of these two paradigms to segmentation-based tracking. 
While previous surveys are based on attributes-based taxonomy, deep network structure and training or introducing a new dataset, we present a comprehensive overview of DCF and Siamese trackers by distinguishing shared as well as specific \textit{open research challenges} in these two popular tracking paradigms. Furthermore, to the best of our knowledge, we are the first to extensively compare the performance of DCF and Siamese trackers on \textit{nine} popular visual tracking benchmarks.  

\section{Discriminative Correlation Filters}
\label{sec:maindcf}
Discriminative correlation filters (DCFs) is a supervised technique for learning a linear regressor. 
In recent years, DCF-based trackers have demonstrated excellent performance on multiple tracking benchmarks. 
The key to the DCF success is the computationally efficient approximation to dense sampling achieved by circularly shifting the training samples, which allows the fast Fourier transform (FFT) to be employed when learning and applying the correlation filter. 
By utilizing the properties of the Fourier transform, a DCF learns a correlation filter online to localize the target object in consecutive frames by efficiently minimizing a least-squares output error. 
In order to estimate the target location in the next frame, the learned filter is then applied to the region of interest in which the location of the maximum response estimates the target location. 
The filter is then updated in an iterative manner by annotating a new sample with this estimate.

\subsection{Standard Single-Channel DCF Formulation}
In the standard single-channel DCF formulation, the aim is to learn a single channel convolution or correlation filter $w$ from a set of training samples $\{(x_{j},y_{j})\}_{j=1}^{m}$.
In its most basic form~\cite{MOSSE10}, each training sample $x_{j}$ consists of a grayscale feature map extracted from an image region.
All samples are assumed to have the same spatial size $N_{1} \times N_{2}$. 
At each spatial location $(n_{1}, n_{2})\in \Omega:= \{0,...,N_{1}-1\} \times \{0,...,N_{2}-1\}$, we thus have an intensity value $x_{j}(n_{1},n_{2})=x_{j}(n)\in \mathbb{R}$, where $n=(n_{1},n_{2})$.
The desired output $y_{j}$ is a scalar valued function over the domain $\Omega$, which contains a label for each location in the sample $x_{j}$.
Typically, $y_{j}$ is set to a sampled Gaussian function with a narrow peak that is centered on the target.

The DCF aims to learn a filter $w$, of the same spatial size $N_{1} \times N_{2}$, such that $x_{j}*w\approx y_{j}$ for all samples $j$ in the dataset. Importantly, the standard DCF formulation employs circular convolution,\footnote{Some authors instead employ circular correlation, obtained by simply reflecting the convolution filter $w$.} 
\begin{equation}
    \label{eq:circconv}
    x * w(n_1,n_2) = \sum_{l_1 = 0}^{N_1-1} \sum_{l_2 = 0}^{N_2-1} x\left((n_1 - l_1)_{N_1}, (n_2 - l_2)_{N_2}\right) w(l_1,l_2).
\end{equation}

It is computed by cyclically shifting the filter $w$ (or equivalently the sample $x$) as ensured by the modulo operation $(a)_N = a \mod N$. Circular convolution \eqref{eq:circconv} can also be seen as first padding the sample $x$ by periodic repetition, followed by regular convolution with the filter $w$.

Since the convolution operation is linear, we can express it in matrix form $x*w =\mathbfcal{C}(x)w$. With slight abuse of notation in order to avoid clutter, $w$ is interpreted as a vectorization of the original filter on the right-hand side of the equation. The matrix $\mathbfcal{C}(x)$ contains a highly regular structure, containing circulant block matrices, where the blocks themselves are also circulant \cite{horn2012matrix}.
Each row in $\mathbfcal{C}(x)$ consists of a cyclicly shifted version of the original patch $x$. The convolution theorem implies that a circulant matrix can be diagonalized as,
\begin{equation}
\mathbfcal{C}(x)=\mathbfcal{F}\textrm{diag}(\hat{x})\mathbfcal{F}^{H}~\textrm{and}~\mathbfcal{C}(x)^{\top}=\mathbfcal{F}\textrm{diag}(\Bar{\hat{x}})\mathbfcal{F}^{H},
\label{eqn1}
\end{equation}

\noindent where $\mathbfcal{F}$ represents the Discrete Fourier Transform (DFT) matrix. 
We denote the conjugate of $x$ by $\bar{x}$ and its Fourier transform $\mathbfcal{F}^{H}x$ by $\hat{x}$, where $^{H}$ denotes the conjugate transpose.

\noindent In order to find the optimal filter $w$, we formulate a linear least-squares regression problem.
For convenience, we first concatenate all samples $j$ in the dataset to get the data matrix $X$ and target vector $\textbf{y}$ as,
\begin{equation}
X = 
\begin{pmatrix}
\mathbfcal{C}(x_{1}) \\
\vdots \\
\mathbfcal{C}(x_{m}) 
\end{pmatrix},
\textbf{y} = 
\begin{pmatrix}
y_{1} \\
\vdots \\
y_{m} 
\end{pmatrix}\,.
\label{eqn2}
\end{equation}
The DCF is learned by minimizing the squared error objective $L(w)$ over the samples of $X$ and their regression targets $\textbf{y}$ as,
\begin{equation}
L(w)=\lvert\lvert Xw-\textbf{y}\rvert\rvert ^{2}+\lambda\lvert\lvert w\rvert\rvert ^{2},
\label{eqn3}
\end{equation}
Here, $\lambda$ is a regularization parameter. As this is a linear least squares problem, the solution is given by the following closed form expression
\begin{equation}
w=(X^{\top}X+\lambda I)^{-1}X^{\top}\textbf{y} \,.
\label{eqn4}
\end{equation}
To compute weights of the filter $w$, we simply substitute (\ref{eqn2}) into (\ref{eqn4})
\begin{equation}
w_{m}=\Big(\sum_{j=1}^{m}\mathbfcal{C}(x_{j})^{H}\mathbfcal{C}(x_{j})+\lambda I\Big)^{-1}\sum_{j=1}^{m}\mathbfcal{C}(x_{j})^{H}y_{j}.
\label{eqn5}
\end{equation}

\begin{figure}[t!]
\centering
\includegraphics[width=3.5in]{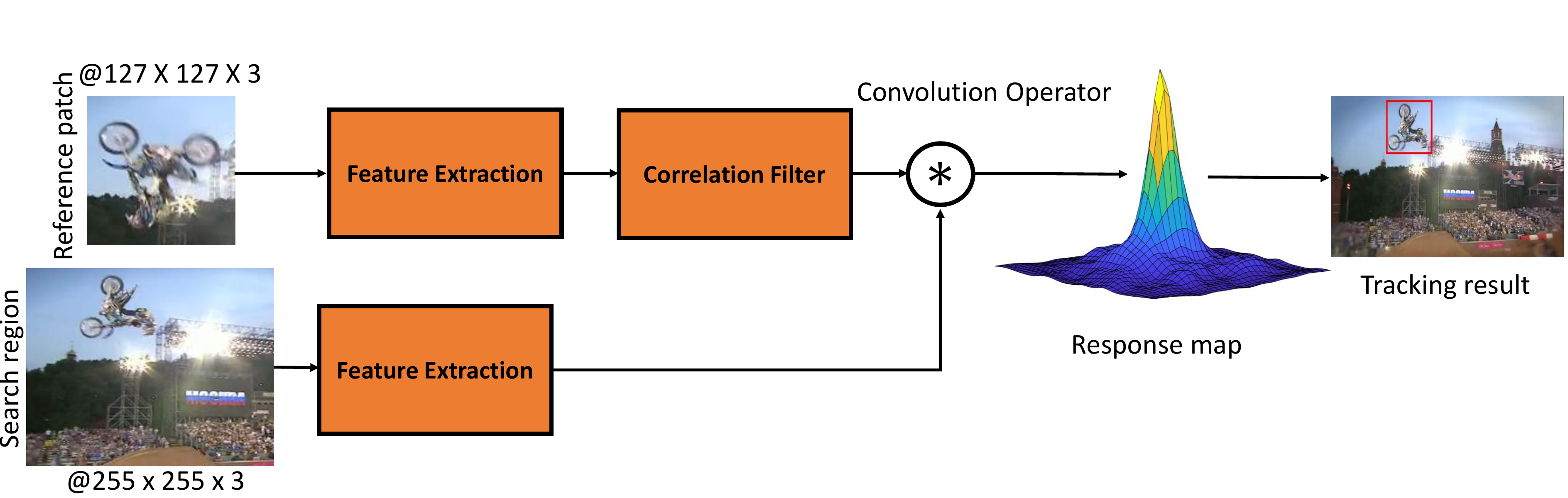}
\vspace{-1.7\baselineskip}
\caption{The tracking pipeline of a classical discriminative correlation filter.}
\label{fig_dcf}
\vspace{-1.3\baselineskip}
\end{figure}

Since the matrix $\mathbfcal{C}(x_{j})$ is circulant, it can be efficiently diagonalized using Eq.~(\ref{eqn1}) as,
\begin{equation}
w_{m}=\mathbfcal{F}\Big(\sum_{j=1}^{m}\textrm{diag}(\hat{x}_{j})^{H}\textrm{diag}(\hat{x}_{j})+\lambda I\Big)^{-1}\sum_{j=1}^{m}\textrm{diag}(\hat{x}_{j})^{H}\hat{y}_{j}.
\label{eqn6}
\end{equation}
Note that the matrix to be inverted in \eqref{eqn6} is diagonal, which can be computed by a simple element-wise division. We can further simplify the expression in terms of its Fourier coefficients of the filter by multiplying with $\mathbfcal{F}^{H}$ from the left, leading to a particularly simple and efficient formula
\begin{equation}
\hat{w}_{m}=\frac{\sum_{j=1}^{m}\bar{\hat{x}}_{j}\hat{y}_{j}}{\sum_{j=1}^{m}\bar{\hat{x}}_{j}\hat{x}_{j}+\lambda}.
\label{eqn7}
\end{equation}
The resulting formula contains only element-wise operations and DFTs. 
It is hence computed in $\mathcal{O}(N \log N)$ time by exploiting the FFT algorithm, where $N=N_{1}N_{2}$ denotes total number of pixels within a sample.

\subsection{Standard Multi-channel DCF Formulation}
In DCFs, multi-channel features, such as the RGB image representation of sample $x_{j}$, offer a much richer representation compared to a single-channel feature (e.g. grayscale intensity). Here, we describe the multi-channel DCFs formulation. In the literature, there are two general strategies to integrate multi-channel features, here termed early and late fusion.

\noindent \textbf{Early Fusion}: The first common scheme is known as early fusion in which the multiple channels are directly concatenated and then a multi-channel DCF classifier is trained.
To generalize \eqref{eqn7} to multi-channel DCFs, we let $x$ be composed of several feature channels $x^d: \Omega \rightarrow \mathbb{R}$, where $d\in \{1,...,D \}$ is the channel index.
In other words, at each spatial location $(n_{1}, n_{2})\in \Omega:= \{0,...,N_{1}-1\} \times \{0,...,N_{2}-1\}$ we have a $D$-dimensional feature vector $x_{j}(n_{1},n_{2})=x_{j}(n)\in \mathbb{R}^{D}$.
Similarly, we extend the filter $w$ to be $D$-dimensional, denoting the individual filter channels with $w^{d}$. 
The convolution between two multi-dimensional functions is defined by simply summing the convolution responses for each individual channel. 
We can learn the filter $w$ by minimizing the generalized objective,
\begin{equation}
L(w)=\sum_{j=1}^{m}\alpha_{j}\left\|\sum_{d=1}^{D}x_{j}^{d}*w^{d}-y_{j}\right\|^{2}+\lambda\sum_{d=1}^{D}\lvert\lvert w^{d} \rvert\rvert^{2}.
\label{eqn9}
\end{equation} 
Here, we have additionally introduced the per-sample importance weights $\alpha_j \geq 0$.

The above multi-channel formulation of DCF is still a linear least squares problem.
It cannot be fully diagonalized by the DFT in the general case. 
There are however two exceptions, where diagonal solutions are found. 
The first case is the original MOSSE model in \eqref{eqn7} and the other case occurs when there is only one single training sample $m = 1$, resulting in the following optimal filter,
\begin{equation}
\hat{w}^{d}=\frac{\Bar{\hat{x}}^{d}\hat{y}}{\sum_{d=1}^{D}\Bar{\hat{x}}^{d}\hat{x}^{d}+\lambda}.
\label{eqn10}
\end{equation}

By again exploiting the FFT, the solution to the general multi-channel objective \eqref{eqn9} can be obtained by solving $N$ number of $D$-dimensional linear systems. This has a computational complexity of $O(DN\log N + ND^{3})$, which is substantially smaller than the direct solution in the spatial domain, which requires $O(D^{3}N^{3})$ operations \cite{KCF14}.

\noindent \textbf{Late Fusion}: Here, the aim is to compute the DCF on each single channel of the sample $x_{k}$ and then aggregate all the filters to obtain the resulting classifier. 
Let $\hat{w}^{d}$ be the estimated filter of the $d$-th feature layer using \eqref{eqn7}, then $w=\sum_{d=1}^{D}\beta_{d}w_{d}$ is aggregated filter obtained by summing up each DCF, where $\beta$ is the scaling factor used to give relative importance to each of the feature layer.

\subsection{Standard DCF Tracking Pipeline}
For tracking, the DCF first learns the filter $w$ online and then it performs tracking-by-detection. 
Once the target is tracked in the current frame, the model is then learnt recursively.
The block diagram of the DCF tracking pipeline is shown in Fig. \ref{fig_dcf}.

\noindent \textbf{Target Detection}: Let $m$ be the number of the current image frame, in which we strive to localize the target. 
From the previous frame we are given the filter $\hat{w}_{m-1}$ that has been recursively updated since the initial frame. 
We extract an image patch $z$ centered at the predicted target location.
Here, $z$ has the same size $N_{1} \times N_{2}$ as the training patches $x_{j}$.
Using the convolution theorem, we then predict target scores $s(n)$ at each location $n \in \Omega$ in $z$ by applying the learned filter,
\begin{equation}
s=\sum_{d=1}^{D}z^{d}*w^{d}_{m-1}=\mathbfcal{F}^{-1}\Big\{\sum_{d=1}^{D} \hat{z}^{d}\hat{w}_{m-1}^{d} \Big\}\,. 
\label{eqn11}
\end{equation}
This computes the score function at all locations $n \in \Omega$.
We can then estimate the target location in frame $m$ as the maximizer of the target score function $n^{*}=\argmax_{n}s(n)$.
In case we are confident about our estimate, we can construct a new training sample $(x_{m},y_{m})$ by extracting a patch $x_{m}$ centered at the estimated location $n^{*}$.

\noindent \textbf{Model update}: Another feature of \eqref{eqn7} is that the filter can be easily updated with new training samples. 
Let us regard the sample index $j$ as the frame number. 
That is, sample $x_{j}$ was extracted from frame number $j$.
In online learning, it is common to employ a learning rate parameter $\gamma \in [0, 1]$ which controls the speed at which the model is adapted to the new data.
Eq. \eqref{eqn7} suggests updating the numerator $\{w^{num}_{m}\}$ and denominator $\{w^{den}_{m}\}$ of the filter $\hat{w}_{m}=\frac{w^{num}_{m}}{w^{den}_{m}}$ recursively as,
\begin{equation}
\begin{split}
\hat{w}_{m}^{num}=(1-\gamma)\hat{w}_{m-1}^{num}+\gamma \Bar{\hat{x}}_{j}\hat{y}_{j},\\
\hat{w}_{m}^{den}=(1-\gamma)\hat{w}_{m-1}^{den}+\gamma (\Bar{\hat{x}}_{j}\hat{x}_{j}+\lambda).
\end{split}
\label{eqn12}
\end{equation}
To achieve the solution given by the recursive formula (\ref{eqn12}), we need to choose the weights $\alpha_{j}=\gamma(1-\gamma)^{m-j}$ for $j>1$ and $\alpha_{1}=(1-\gamma)^{m-1}$ in \eqref{eqn9}.

\subsection{\textbf{Open Issues in the Standard DCF Tracking Pipeline}}
While having important promising properties, the standard DCF framework presents several challenges including feature representations,
boundary artifacts, optimization, and target state estimation when applied for the task of generic object tracking. 
In the subsections below, we identify and discuss these important challenges developing a DCF based tracking pipeline.

\subsubsection{\textbf{Feature Representations}}
\label{sec:feautures}
In object tracking, a variety of visual features have been investigated in the literature. 
Finding discriminative, yet invariant features is particularly important when applying linear discriminative models, such as DCF, which are restricted to finding a linear decision boundary. 
Handcrafted features \cite{dalal2005histograms, felzenszwalb2009object, van2009learning}, deep features \cite{simonyan2014very, chatfield2014return, he2016deep}, hybrid features, and end-to-end learning of features \cite{ATOM19} have been explored within the DCF-based tracking framework. 
Next, we describe details about different types of features used in the DCF-based trackers.

\noindent \textbf{Handcrafted Features:}
Early DCF trackers such as, MOSSE \cite{MOSSE10} and CSK \cite{CSK12} have exploited intensity features for object tracking. Other than intensity features, local color and intensity histograms features are also utilized in DCF trackers such as, RPAC \cite{RPAC15}, LCT+ \cite{LCT+18}, LCT \cite{LCT15}, and CACF \cite{CACF17}.
Simple color representations, including RGB and LAB have been used for DCFs-based trackers such as, STAPLE \cite{STAPLE16} (RGB), SCT \cite{SCT16} (RGB $+$ LAB), and ACFN \cite{ACFN17} (RGB $+$ LAB).

To achieve a more discriminative image representation, ACA \cite{ACA14} investigated different color descriptors and proposed to use the Color Names (CN) features \cite{van2009learning} along with the intensity channel.
The ACA tracker further introduced an adaptive dimensionality reduction technique to compress the CN features, thereby providing a tradeoff between speed and tracking performance. 
CN features have also been employed in several subsequent DCFs-based trackers such as, MCCT \cite{MCCT18}, MKCF \cite{MKCF15}, MUSTer \cite{MUST15}, CSR-DCF \cite{CSRDCF17}, CCOT \cite{CCOT16}, ECO \cite{ECO17}, UPDT \cite{UPDT18}, AutoTrack \cite{AUTO20}, ARCF \cite{ARCF19}, GFS-DCF \cite{GFSDCF19}, RPCF \cite{RPCF19}, and DRT \cite{DRT18}.

Another popular handcrafted feature employed in DCF-based trackers is Histogram of Oriented Gradients (HOG) \cite{dalal2005histograms}. 
HOG captures shape information by collecting statistics of the image gradients. 
HOGs are formed in a dense image grid of cells.
Within the DCF paradigm, KCF~\cite{KCF15} was the first tracker to utilize HOG features for tracking. 
Several DCF trackers such as, MCCF \cite{MCCF13}, CFLB \cite{CFLB15}, BACF \cite{BACF17}, SRDCF \cite{SRDCF15}, STRCF \cite{STRCF18}, RPCF \cite{RPCF19}, GFS-DCF \cite{GFSDCF19}, RPT \cite{RPT15}, RCF \cite{RCF16}, LMCF \cite{LMCF17}, PTAV \cite{PTAV17}, StruckCF \cite{StruckCF16}, CFAT \cite{CFAT16}, and LSART \cite{LSART18} have utilized HOG features. 
\textit{These features have been the preferred alternative among handcrafted methods due to its speed and effectiveness. Further, HOG features have also been effectively combined with CN features to utilize both shape and color information.}

\noindent \textbf{Deep Features:}
In recent years, deep learning has revolutionized many areas in computer vision.
Deep Convolutional Neural Networks (CNNs) have shown to be particularly well suited for image related tasks \cite{goodfellow2016deep}. They apply a sequence of learnable convolutions and non-linear operations onto the image. 
However, employing deep features for object tracking has proved challenging.
This was primarily due to the scarcity of training data for tracking in the initial years of deep learning, as well as the high dimensionality of the features. 
Therefore, many DCF-based trackers such as, HCF \cite{HCF15}, HDT \cite{HDT16}, CCOT \cite{CCOT16}, ECO \cite{ECO17}, ASRCF \cite{ASRCF19}, and RPCF \cite{RPCF19}, employ deep CNNs that are pre-trained on the ImageNet dataset \cite{deng2009imagenet} for image classification.
Despite being trained for classification, such deep representations are applicable to a wide range of vision tasks \cite{sharif2014cnn}. 

Within the DCF framework, the deep features are extracted from the convolutional layers. 
Some popular pre-trained deep networks, including VGG-19 \cite{simonyan2014very}, imagenet-vgg-m-2048 \cite{chatfield2014return}, VGG-16 \cite{simonyan2014very}, ResNet50 \cite{he2016deep}, and GoogleNet \cite{szegedy2015going} are used to extract deep feature representations. 
For instance, Ma \textit{et al.,} incorporated the hierarchical deep convolutional features for visual tracking \cite{HCF15}. 
The HDT tracker also employed deep features from six convolutional layers of the same network \cite{HDT16}. 
The DeepSRDCF tracker used the imagenet-vgg-m-2048 network and performed an analysis of the convolutional feature maps for tracking \cite{DEEPSRDCF15}, indicating the importance of the shallow layer. 
Shallow layers contain low-level information at a high spatial resolution, important for accurate target localization.

On the other hand, deeper-layer feature maps possess high-level invariance to complex appearance changes, such as deformations and out-of-plane rotations. 
Thus, deeper layers have the potential of improving tracking robustness, while largely invariant to small translation and scale changes. 
Therefore, an accurate strategy of fusing shallow and deeper convolutional layers within the DCF framework has been a topic of interest. 
In CCOT, a continuous-domain formulation of the DCF framework is proposed that enables the integration of multi-resolution features \cite{CCOT16}.
The ECO investigates strategies to reduce the computational cost of the CCOT and mitigates the risk of overfitting \cite{ECO17}.
Other trackers, such as HDT \cite{HDT16}, HCFTs \cite{HCFTs19}, MCCT \cite{MCCT18}, MCPF \cite{MCPF17}, MCPFs \cite{MCPFs18}, LMCF \cite{LMCF17}, STRCF \cite{STRCF18}, TRACA \cite{TRACA18}, DRT \cite{DRT18}, UPDT \cite{UPDT18}, and GFS-DCF \cite{GFSDCF19} integrate deep features using late fusion strategy. This strategy is to train a classifier on each individual feature representation and then aggregate feature response map. 

\noindent \textbf{End-to-End Features Learning:}
This scheme within the DCF framework has also been explored. 
Here, the backbone networks, such as AlexNet \cite{krizhevsky2017imagenet}, VGG-16 \cite{simonyan2014very} or ResNet50 \cite{he2016deep}, are used to optimize deep features on tracking datasets \cite{SACF18, CFNET17, ACFN17}. Instead of relying on pre-trained networks, the task-specific deep feature learning promises improved representations for the tracking problem itself.
Valmadre \textit{et al.} proposed a CFNET that follows end-to-end learning of correlation filters in an offline manner \cite{CFNET17}. 
Other trackers such as, CREST \cite{CREST17} and ACFN \cite{ACFN17} also followed the same strategy in an online manner. 
In these trackers, the objective is to improve the target regression.
These methods demonstrated comparable performance, as compared to trackers that utilized deep features extracted from the pre-trained networks. 
Recently, an end-to-end target scale estimation component is incorporated in ATOM \cite{ATOM19} while the discriminative strength of the classical DCF model is improved in DiMP \cite{DIMP19} and PrDiMP \cite{PRDIMP20}. 
\textit{This recent trend of end-to-end features learning in DCF trackers \cite{ATOM19, DIMP19, PRDIMP20} has resulted in excellent tracking performance on multiple benchmarks, paving the way to explore more sophisticated end-to-end feature learning in DCF paradigm.}

\subsubsection{\textbf{Boundary Artifacts}}
\label{sec:boundaryeffects}
The standard convolution is effectively replaced by circular convolution \eqref{eq:circconv} in the DCF formulation to ensure the applicability of the DFT, resulting in the formula \eqref{eqn11} for evaluating the target predictions. 
This change may seem minor and might even have gone unnoticed to some readers. 
However, the circular convolution introduces unwanted boundary artifacts.

The most severe consequence of the circular convolution assumption arises in the learning formulation \eqref{eqn9}.
The fundamental notion of the DCF paradigm is to train a filter $w$ that can discriminate the target from background image regions. 
Due to the periodic effects, most of the original background content is replaced by synthetic repetition of a smaller image patch.
The model thus sees fewer background samples during training, severely limiting its discriminative power. 
Furthermore, due to the distortions caused by the periodic repetitions, the predicted target scores are only accurate near the center of the image patch. 
The size of the search area is therefore limited. 
As traditionally performed in signal processing, DCF methods typically pre-process the samples $x$ by multiplying them with a window function \cite{MOSSE10}.
However, this technique does not attempt to solve the aforementioned problems and only serves to smooth out the discontinuities at border regions.
Several solutions have been proposed in the literature to overcome the aforementioned boundary artifacts problems. 
To this end, several approaches are proposed that incorporate target-specific spatial, spatiotemporal, and smoothness constraints within the DCF objective function \cite{CFLB15, SRDCF15, BACF17, STRCF18, ASRCF19}. 
Below, we summarize some of these major developments. 

\noindent \textbf{Spatial regularization:}
In the SRDCF, Danelljan \textit{et al.} proposed a spatially regularized framework to control the spatial extend of the filter in order to alleviate the boundary problem \cite{SRDCF15}. 
A spatial regularization component is integrated into the multi-channel DCF formulation \eqref{eqn9} as,
\begin{equation}
L(w)=\sum_{j=1}^{m}\alpha_{j}\lvert\lvert \sum_{d=1}^{D}x_{j}^{d}*w^{d}-y_{j}\rvert\rvert^{2}+\lambda\sum_{d=1}^{D}\lvert\lvert fw^{d} \rvert\rvert^{2},
\label{eqn14}
\end{equation} 
The spatial weight function $f: \Omega \rightarrow \mathbb{R}$ takes positive values $f(n)>0$, penalizes the filter coefficients $w^{d}(n)$ based on their spatial location $n$.
By letting $f$ takes large values at background pixels and small values inside the target region, background filter coefficients are penalized. 
As a result, a compact filter $w$ focusing on the target region can be learned, even for large image sample sizes. The spatial regularization strategy has been employed in a variety of trackers, including ARCF \cite{ARCF19}, ASRCF \cite{ASRCF19}, and AutoTrack \cite{AUTO20}.
In order to improve the efficiency of the SRDCF formulation, Li \textit{et al.} proposed the STRCF \cite{STRCF18} which only employs a single training sample, and instead introduces a temporal regularization term to integrate historic information. 
It has also become a popular baseline for numerous works \cite{AUTO20}.

\noindent \textbf{Constraint optimization:} 
While the SRDCF~\cite{SRDCF15} aims to penalize the filter coefficients outside the target region, Kiani \textit{et al.} propose to introduce hard constraints \cite{CFLB15,BACF17}.
This strategy enforces that the filter coefficients $w(n)$ are zero outside the target region. The resulting DCF formulation can be expressed as follows by introducing a binary mask $P$, 
\begin{equation}
L(w)=\frac{1}{2}\sum_{j=1}^{m}\lvert\lvert y_{j}-(P \cdot w) * x_{j}\rvert\rvert^{2}_{2}+\frac{\lambda}{2}\lvert\lvert w \rvert\rvert^{2}_{2},
\label{eqn13}
\end{equation}

\noindent Here, the element-wise product $P \cdot w$ effectively masks out the influence of filter weights corresponding to background features.
The resulting optimization problem can be efficiently solved through iterative techniques such as ADMM \cite{CFLB15,BACF17}.
The aforementioned target-specific constraints investigated in \eqref{eqn14} and \eqref{eqn13} are usually fixed for different objects and they do not vary during the tracking process.
Recently, Dai \textit{et al.} extended BACF and SRDCF by introducing an adaptive regularization term \cite{ASRCF19}.

\noindent \textbf{Implicit methods:}
With the GFS-DCF, Xu \textit{et al.} proposed a joint group feature selection model that simultaneously learns three regularization terms including spatial regularization for feature selection, channel regularization for feature channel selection, and low-rank temporal regularization term to enforce smoothness on the filter weights \cite{GFSDCF19}.
Mueller \textit{et al.} proposed to regularize the contextual information of each target patch \cite{CACF17}. 
In every frame, CACF samples several context patches, which serve as negative samples. 

\noindent \textbf{Spatial formulation:} 
Danelljan \textit{et al.} and Bhat \textit{et al.} proposed ATOM \cite{ATOM19} and DiMP \cite{DIMP19} trackers. 
Both these trackers employ deep features in low resolution (stride 16) in order to first coarsely, yet robustly, localize the target object. Due to the coarse resolution and the resulting small target filter size ($4 \times 4$), Danelljan \textit{et al.} found that the the filter can be directly learned in the spatial domain using dedicated and efficient iterative solvers \cite{ATOM19}.
This approach allows both ATOM and DiMP to completely circumvent the boundary artifacts problem, since no periodic extension of the training samples are performed.

Both the regularization-based (SRDCF \cite{SRDCF15}, STRCF \cite{STRCF18}) and constraint-based (CFLB \cite{CFLB15}/BACF \cite{BACF17}) formulations have seen large success and been employed in a wide range of trackers.
However, more recent deep learning methods (ATOM \cite{ATOM19}/DiMP \cite{DIMP19}), have completely circumvented the boundary artifact problem by directly optimizing the filter in the spatial domain. Thus, while the Fourier domain is computationally attractive for high-resolution feature maps, efficient spatial-domain optimization methods prevail for online learning when using powerful low-resolution deep features. \textit{Recent works constituting the current SOTA in DCF-based tracking \cite{DIMP19,PRDIMP20} therefore employ a purely spatial formulation, which does not require additional strategies for alleviating boundary effects. By further extending the filter to a multi-channel output, the latter strategy has also demonstrated its use for segmentation \cite{bhat2020learning}.}

\subsubsection{\textbf{Optimization}}
\label{sec:optimization}
In the standard DCF formulation, inference is performed by computing the DFT coefficients $\hat{w}$ using the least squares solution \eqref{eqn7}. 
However, when the model grows more complex and advanced e.g., by introducing multiresolution feature maps and target-specific constraints, such as spatial regularization, and temporal regularizations, the model inference cannot be performed using the simple least-square solution, as provided in \eqref{eqn13}. 
Since computational efficiency is a crucial factor in most applications, these modifications require alternative inference methods. Therefore, finding efficient and robust inference schemes is a key problem in DCF-based tracking.

Model inference is performed by minimizing the multi-channel loss \eqref{eqn9}, which forms the basis of the DCF framework. 
However, it does not allow for any efficient closed-form solution. Therefore, many DCF trackers such as CACF \cite{CACF17}, CSK \cite{CSK12}, KCF \cite{KCF15}, MUSTer \cite{MUST15}, SP-KCF \cite{SPKCF17}, and CFAT \cite{CFAT16} employ the diagonalizable cases in both primal or dual domains to derive approximate model inference schemes. 
These loss functions rely on the very restrictive assumptions of a single feature channel $D = 1$ and a single training sample $m = 1$, respectively. 
Moreover, these solutions cannot benefit from the aforementioned additional regularizations. Several efficient optimization methods for model inference have been introduced to minimize the loss functions in the literature.

\noindent \textbf{Gauss-Seidel Method \cite{hadjidimos2000successive}:} Minimizing the DCF loss functions with its spatially regularized variants online is a highly challenging problem, since the filter $w$ contains tens or hundreds of thousand parameters to be optimized.
In \cite{SRDCF15}, an optimization approach based on the iterative Gauss-Seidel method is proposed to minimize the spatially regularized loss function \eqref{eqn14}.
The same strategy is also considered in its variant DeepSRDCF \cite{DEEPSRDCF15} that employs deep features.
By employing the Gauss-Seidel-based optimization, the tracker achieves a tracking speed of few frames per second.
While not yet real-time, it demonstrated superior robustness and accuracy compared to previous approaches, yet faster than many of its competitors.

\noindent \textbf{Conjugate Gradient Based Method \cite{nocedal2006numerical}:}
To pave the way for the use of deep features and to further improve the computational efficiency, a Conjugate Gradient (CG)-based strategy is utilized in CCOT \cite{CCOT16}.
CG can be applied to any set of normal equations $A\widetilde{w} = b$ of full rank. 
It is based on finding a set of conjugate directions $p^{(i)}$ and optimal step lengths $\beta^{(i)}$ used for updating the filter $\widetilde{w}^{(i)}=\widetilde{w}^{(i-1)}+\beta^{(i)}p^{(i)}$.
Theoretically, the algorithm converges to the solution in a finite number of iterations $i$. 
In practice however, the algorithm is stopped after a fixed number of iterations or when the error has decreased to a satisfactory level.

The computational bottleneck of CG is the evaluation of the matrix-vector product $Ap^{(i)}$ in each iteration $i$. 
This is in fact the key advantage of CG. 
The Gauss-Seidel method requires solving a triangular system in each iteration. 
On the contrary, CG can exploit the particular sparsity structure of $A$ in the computation $Ap^{(i)}$.
Moreover, $Ap^{(i)}$ can be implemented as a series of simple block-wise dense matrix-vector products, convolutions and point-wise multiplications. 
This reduces the quadratic $O(D^{2})$ complexity in the feature dimension to linear $O(D)$, a crucial improvement enabling tractable integration of high-dimensional deep features. 

To address non-linear least squares problems, the Gauss-Newton optimization is also used in many trackers, including ECO \cite{ECO17}, ATOM \cite{ATOM19}, and UPDT \cite{UPDT18}.
The method linearizes the error residuals using a Taylor series expansion around the current estimate to find a quadratic approximation of the objective. The resulting quadratic problem can then be tackled with iterative methods, for instance the CG approach described above. 
In the ECO~\cite{ECO17} and ATOM~\cite{ATOM19}, the Gauss Newton combined with CG is employed to jointly optimize the filter $w$ and the dimensionality reduction matrix. 
DiMP employs Gauss Newton together with Steepest Descent iterations to learn the filter $w$ using a non-linear robust loss function \cite{DIMP19}.
The optimization steps are themselves differentiable, which further enables end-to-end learning of the underlying deep features. 
The PrDiMP further employs the more general Newton approximation to address the convex and non-linear KL-divergence objective \cite{PRDIMP20}.

\noindent \textbf{Alternating Direction Method of Multipliers (ADMM) Method \cite{boyd2011distributed}:} When a classical DCF formulation grows in terms of additional regularizations such as those presented in models \eqref{eqn14}-\eqref{eqn13}, the formulation becomes a constraint optimization problem which can be solved using efficient convex or non-convex solvers such as an ADMM-based optimization method. 
The ADMM method has recently been used in many DCF-based trackers to efficiently solve the DCF loss functions specially when additional regularizations are introduced. 
The ADMM-based optimization approach provides closed-form solutions for each sub-problem and empirically converges within a very few iterations. 
In ADMM, the model is solved by breaking it into smaller pieces, each of which is then easier to handle. 
The augmented lagrangian formulation is always used to convert the constrained optimization model into unconstrained model with a lagrangian penalty as an additional variable. 
Each sub-problem of the unconstrained model is then solved in an iterative manners.
Trackers such as BACF \cite{BACF17}, DRT \cite{DRT18}, AutoTrack \cite{AUTO20}, ARCF \cite{ARCF19}, and RPCF \cite{RPCF19} have employed ADMM for efficient solutions.
Other optimization methods such as Gradient descent is used by ACFN \cite{ACFN17}, CREST \cite{CREST17}, and DSLT+ \cite{DSLT+20} trackers for network optimization.

\noindent \textit{\noindent \textbf{Current Choices:} Both Conjugate Gradient \cite{ECO17,CCOT16,ATOM19,UPDT18} and ADMM  (STRCF \cite{STRCF18}, BACF \cite{BACF17}, RPCF \cite{RPCF19}, ASRCF \cite{ASRCF19}, GFS-DCF \cite{GFSDCF19}, and MCCT \cite{MCCT18}) have been popular choices for DCF trackers.
Both support the more complex DCF formulations required to mitigate the boundary artifacts, using e.g.\ spatial regularization or constraints.
The most recent approaches DiMP employed a steepest descent in a spatial-domain formulation \cite{DIMP19}, while PrDiMP combined it with a Newton approximation to minimize the non-linear KL-divergence objective \cite{PRDIMP20}.
In addition to efficiency, the adoption of end-to-end learning have brought the importance of differentiability and simplicity of the optimizer, motivating the latter choice. It has also opened the opportunity to train components and hyper-parameters of the optimizer in a meta-learning fashion, setting the stage for future research directions.}

\subsubsection{\textbf{Target State Estimation}}
\label{sec:stateestimation}
The DCF models have demonstrated promising results in terms of accuracy and robustness.
However, when a target moves its template size (also known as bounding box size) also changes.
The DCF model employs the bounding box with the fixed size that often leads to severe model drift, since it is not capable to handle target scale variations.
Accurate scale estimation poses a great challenge in the classical DCF formulation. 
The problem of handling bounding box size for accurate target scale estimation is an established research direction. 
Here, we briefly describe the most popular strategies.

\noindent \textbf{Multiple Resolution Scale Search Method:} One straight-forward strategy is to apply learned translation filter $w$ at different image scales. 
That is, the image is first resized by different scale factors, followed by feature extraction. 
The feature map at each scale is then convolved with learned filter $w$ to compute the target scores. 
The change in target location and scale can then be estimated by finding maximum score across all scales. 
This is a common strategy, often applied in both tracking and detection \cite{felzenszwalb2009object}.  

Li \textit{et al.} proposed SAMF tracker in which the translation and scale filters were jointly trained using the standard DCF formulation \cite{SAMF14}.
The results demonstrated significant performance gain compared to the standard DCF. 
This scale adaptive component has been utilized in a number of DCF-based trackers such as, CACF \cite{CACF17}, CFAT \cite{CFAT16}, and FD-KCF \cite{FDKCF19}.
However, this approach suffers from a higher computational cost, since the translation filter has to be applied at several resolutions to achieve sufficient scale accuracy.

\noindent \textbf{Discriminative Scale Space Search Method:}
Danelljan \textit{et al.} proposed an alternative strategy for accurate scale estimation \cite{DSST14}.
Unlike \cite{SAMF14}, the target estimation is performed in two steps in order to avoid an exhaustive search across translation and scales.
Since the scale changes between two frames is usually small or moderate, the target translation is first found by applying the normal translation filter $w$ at the current scale estimate.
Then, a separate one-dimensional filter is applied in the scale dimension to update the target size. 
The scale filter is trained analogously to the translation filter, but operates in the scale dimension by extracting samples of the target appearance from a set of different scales.

The advantage of the aforementioned scale-filter approach \cite{DSST14} is two-fold. 
First, computational efficiency is gained by reducing the search space. 
Second, the scale filter is trained to discriminate between the appearance of the target at different scales, which can lead to a more accurate  estimation.
The proposed scale filter component has been utilized in a multitude of trackers, including STAPLE \cite{STAPLE16}, MUSTer \cite{MUST15}, ASRCF \cite{ASRCF19}, CACF \cite{CACF17}, BACF \cite{BACF17}, CSR-DCF \cite{CSRDCF17}, MCCT \cite{MCCT18}, and LCT \cite{LCT15}.
Moreover, the follow-up fDSST tracker reduces the computational cost of DSST by applying the PCA and sub-grid interpolation \cite{fDSST16}.

\noindent \textbf{Deep Bounding Box Regression Method:}
The aforementioned methods show improved performance. However, they depend on the scaling factor parameters and online accurate correlation filter response. 
These methods do not exploit the powerful deep feature representations in an offline manner. 
Moreover, these online methods do not perform any bounding box regression. As a result, these methods show performance degradation in the presence of sudden scale variations.
Accurate estimation of the target object bounding box is a complex task, requiring high-level a priori knowledge. 
The bounding box depends on the pose and viewpoint of the target, which cannot be modeled as a simple image transformation (e.g.\ uniform image scaling). 
It is therefore extremely difficult to learn accurate target estimation online from scratch.

In object detection methods, bounding box regression has been widely employed for precise localization of object bounding boxes \cite{girshick2015fast, jiang2018acquisition, girshick2014rich, huang2017speed, redmon2017yolo9000, redmon2016you}. 
Conventional $L_{1}$ or $L_{2}$ loss is used between the predicted and ground-truth parameters of the bounding box for regression. 
To exploit the strength of end-to-end deep features learning for target scale estimation, this component has recenty been utilized in \cite{ATOM19, DIMP19}.
In ATOM \cite{ATOM19}, inspired by the IoU-Net \cite{jiang2018acquisition}, the targets-specific features are trained.
Since the original IoU-Net is class-specific, and hence not suitable for generic tracking, a novel architecture is proposed for integrating target-specific information into the IoU prediction \cite{ATOM19}. 
This is achieved by introducing a modulation-based network component that incorporates the target appearance in the reference image to obtain target-specific IoU estimates. 
This further enables the target estimation component to be trained offline on large-scale datasets. 
During tracking, the target bounding box is found by simply maximizing the predicted IoU overlap in each frame.
Results demonstrated an excellent performance boost as compared to classical multi-scale search methods.

Several recent trackers, including DiMP \cite{DIMP19}, PrDiMP \cite{PRDIMP20}, and KYS \cite{KYS20} have also utilized this strategy for state estimation.
In PrDiMP, instead of predicting the IoU, it employs energy-based model that predicts the un-normalized probability density of the bounding box. 
This is trained by minimizing the KL-divergence to a Gaussian model of the label noise. Fig. \ref{fig2_infl} shows  influential DCF trackers in literature. 

\subsection{Evolution of DCFs to Segmentation-based Trackers}
Precise and accurate object segmentation provides reliable object observations for tracking, which can solve several tracking problems including rotated bounding box issues, occlusion, deformation, and scaling, etc., and fundamentally avoid tracking failures. 
In literature, segmentation-based approaches have been incorporated within the DCFs-based trackers for improved filter learning in the presence of non-rectangular targets \cite{CSRDCF17, OTR19, STAPLE16}.

Bertinetto \textit{et al.}, used a color histogram-based segmentation method to improve tracking under varying illumination changes, motion blur, and target deformation \cite{STAPLE16}.
Lukezic \textit{et al.}, proposed a spatial reliability map using a color-based segmentation method to regularize filter learning \cite{CSRDCF17}.
A real-time tracker is proposed using handcrafted features and achieved comparable performance using deep features. 
Kart \textit{et al.} \cite{OTR19}, extended the CSR-DCF tracker for RGB-depth tracking based on both color and depth segmentation as depth cues provide more reliable segmentation map.
Lukezic \textit{et al.}, proposed a single shot segmentation tracker to address VOT and video object segmentation problems within a joint framework \cite{D3S20}.
The target is encoded with two discriminative models for the joint tracking and segmentation task. 
The results are reported in many tracking and segmentation benchmarks and demonstrated the benefits.
Recently, Robinson \textit{et al.,} employed a powerful discriminative model using the fast optmization scheme borrowed from the ATOM \cite{ATOM19} for video object segmentation task \cite{robinson2020learning}.
Bhat \textit{et al.,} also used the target model discriminative capabilites for more robust video object segmentation \cite{bhat2020learning}.

\section{Siamese Trackers}
\label{sec:mainsiamese}
Deep learning models have revolutionized many machine learning applications. 
The key to the success of these models is the offline learning capabilities of features on a large volume of data. 
Such offline training models have a capability to learn complex and rich relationships from large amount of annotated data.

End-to-end offline training models have also been employed in the generic object tracking by posing it as a similarity learning problem \cite{SIAMFC16, GOTURN16, SINT16}. 
Deep SNs have been widely used to learn a similarity between the target image and the search image region \cite{SIAMFC16}. 
SNs were first used for signature verification task \cite{bromley1993signature} and then adapted for other applications including fingerprint recognition \cite{chopra2005learning, taigman2014deepface}, stereo matching \cite{zbontar2015computing}, ground-to-aerial image matching \cite{lin2015learning}, and local patch descriptor learning \cite{han2015matchnet}.

\begin{figure}[t!]
\centering
\includegraphics[width=\linewidth]{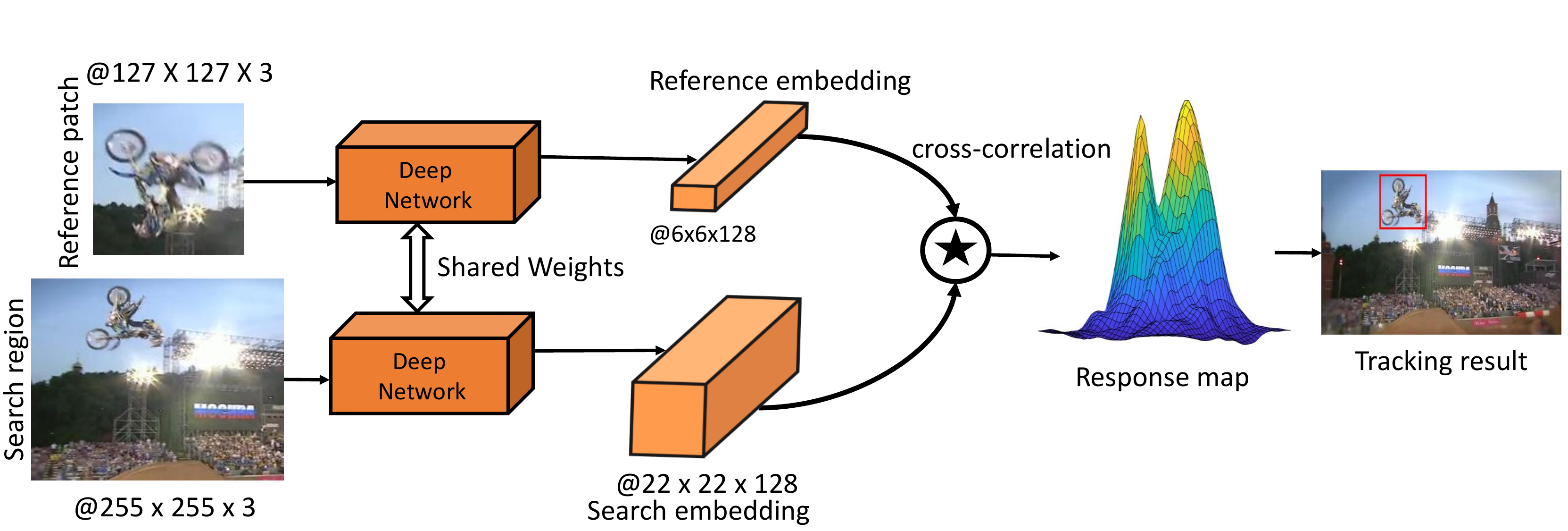}
\vspace{-1.7\baselineskip}
\caption{The Siamese tracking pipeline for generic object tracking.}
\label{fig_siamese}
\vspace{-1.3\baselineskip}
\end{figure}

In VOT, an offline deep network is trained on a large amount of pairs of target images to learn a matching function during training and then this network as a function is evaluated online during tracking.
Bertinetto \textit{et al.} unveiled the power of SNs for VOT \cite{SIAMFC16}.
The siamese tracker consists of two branches, the template branch and the detection branch. 
The template branch receives the target image patch in the previous frame as input while the detection branch receives the target image patch in the current frame as input. 
Both of these branches share CNN parameters so that the two image patches encode the same transformation which is suitable for tracking.
Fig. \ref{fig_siamese} presents the tracking pipeline of a standard SN.
The main aim of SN is to overcome the limitations of pre-trained deep CNNs and take full advantage of end-to-end learning for real-time applications.
The offline training videos are used to instruct the tracker to handle rotations, changes in viewpoint, lighting changes, and other complex challenges.
With the use of a SN, the tracker is able to learn the generic relationship between the object motion and appearance and can be used to locate unseen targets which were not utilized in training.

\begin{figure*}[t!]
\centering
\includegraphics[width=\linewidth]{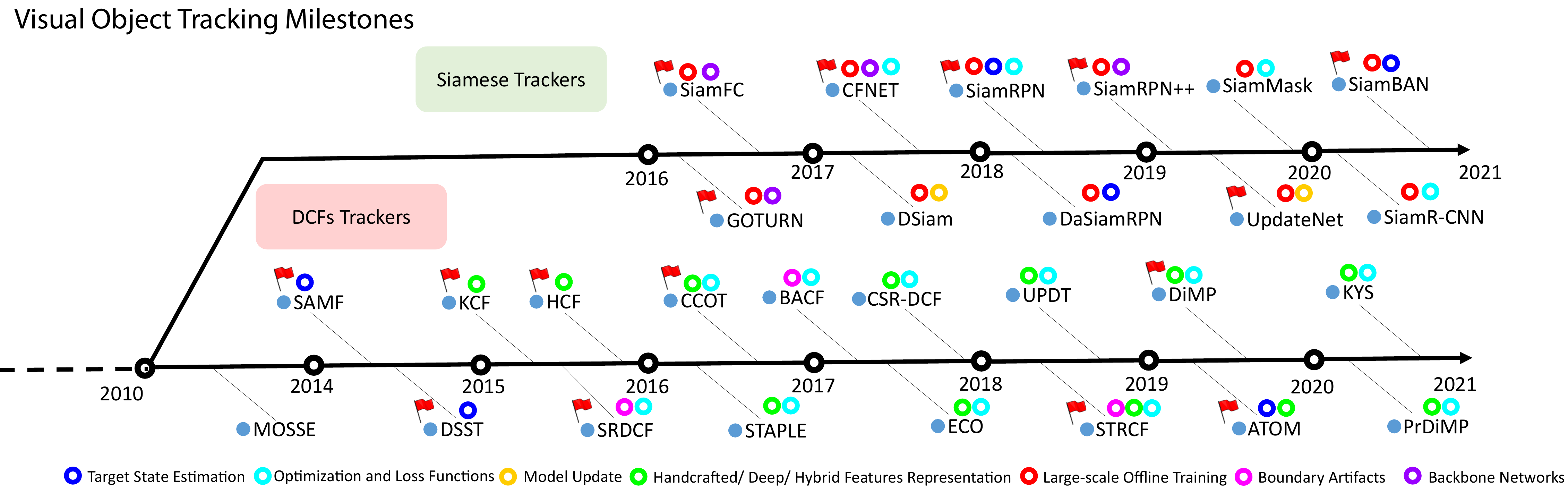}
\vspace{-1.7\baselineskip}
\caption{Influential Siamese and DCF trackers proposed in the 2014-2020 period.}
\label{fig2_infl}
\vspace{-1.3\baselineskip}
\end{figure*}

\noindent \textbf{Training Pipeline:} In the Fully Convolutional SN (SiamFC) \cite{SIAMFC16}, we consider a pairs of training images $(x, z)$. Here, $x$ denotes the object of interest (for instance an image patch cropped from the center of the target image in the first frame) and testing image $z$ represents a larger search area in the next frame. 
We input these pairs $(x,z)$ into a CNN to obtain two feature maps, which are then matched using a cross-correlation,

\begin{equation}
g_{\rho}(x,z)=f_{\rho}(x)\star f_{\rho}(z)+b, 
\label{eqn20}
\end{equation}
 
\noindent where $\star$ denotes cross-correlation operation, $f_{\rho}(.)$ is a deep CNN, e.g.\ AlexNet \cite{krizhevsky2017imagenet}, with learnable parameters $\rho$, and $g_{\rho}(x,z)$ is a response map denoting the similarity between $x$ and $z$, and $b \in \mathbb{R}$ denotes a scalar offset value.
The cross-correlation similarity function \eqref{eqn20} performs an exhaustive search of the features extracted from sample $x$ over the features extracted from $z$.

The goal is for the maximum value of the response map $g_{\rho}(x,z)$  to correspond to the target location.
To achieve it, the network is trained offline on millions of random pairs $(x_{i}, z_{i})$ extracted from a collection of videos to track the generic objects.
The mean of the logistic loss is typically employed to train the network as,

\begin{equation}
\ell(c,v)=\frac{1}{N}\sum_{i =1}^N\log(1+\exp(-c_{i}v_{i})),
\label{eqn21}
\end{equation}

\noindent where $v$ represents real-valued score of a single target-test sample and labels $c_{i}\in\{-1,1\}$, with the true object location belonging to the positive class and all others to the negative class. 
Training proceeds by minimizing an element-wise logistic loss $\ell$ over the training set as:

\begin{equation}
\argmin_{\rho} \mathbb{E}_{(x,z,c)}\ell(c_{i}, g_{\rho}(x_{i},z_{i};\rho)).   
\label{eqn23}
\end{equation}

\noindent Tao \textit{et al.} proposed SINT in which Euclidean distance is employed as a similarity measure instead of cross-correlation \cite{SINT16}.
Held \textit{et al.} proposed the GOTURN in which a bounding box regression is employed \cite{GOTURN16}.
Similarly, Valmadre \textit{et al.} proposed CFNET in which correlation filter is added in the $x$ as a separate block in the matching function \eqref{eqn20} and makes this network shallower but more efficient \cite{CFNET17}.

\noindent \textbf{Testing Pipeline:} The network only provides a function to measure the similarity of two image patches. 
It is necessary to combine this network for VOT with a procedure that describes the logic.
To assess the utility of the similarity function, a simplistic tracking algorithm is employed by first extracting the feature representation of the $x$ and $z$ in the new frame. 
The feature representation of $x$ is then compared to that of $z$, which is obtained in each new frame by extracting a window centred at the previously estimated position, with an area that is four times the size of the object. 

The new position of the target is predicted using the location with the highest score.
The original SiamFC network simply compares every frame to the initial appearance of the object and tracks the object real time at 140FPS on a GPU.
SNs are computationally efficient in both inference and in offline learning. 
SNs have demonstrated SOTA tracking performance and therefore siamese trackers are receiving a lot of attention in the tracking community nowadays.

\subsection{\textbf{Open Issues in Standard Siamese Tracking Pipeline}}
The classical SN outperforms DCFs trackers in both accuracy and efficiency.
However, SN also suffers from several limitations in terms of the backbone features extraction network, need for a large annotated image pairs in offline training, lack of online adaptability, loss function formulation, and the target state estimation. In the subsections below, we  identify and discuss these important challenges for developing a robust Siamese-based tracker.
Below, we briefly describe the details of these problems in siamese tracking and their potential solutions developed in recent years.

\subsubsection{\textbf{Backbone Architectures}}
\label{sec:backbone}
In offline training, the backbone feature extraction network plays a dominant role to capture high-level semantic information of the target. 
Siamese trackers have demonstrated encouraging performance using powerful backbone networks. 
In early siamese trackers (SiamFC \cite{SIAMFC16}, GOTURN \cite{GOTURN16} and SINT \cite{SINT16}), a modified pre-trained AlexNet is fine-tuned \cite{krizhevsky2017imagenet}. 
A variety of trackers (FlowTrack \cite{FLOWTRACK18}, MemTrack \cite{MEMTRACK18}, EAST \cite{EAST17}, and SiamRPN \cite{SIAMRPN18}) have used AlexNet.
However, it is observed that these trackers are still limited in performance because AlexNet is a relatively shallow network and does not produce very strong feature representations.

With the advent of wider and modern deeper networks, the tracking community has also designed SNs based on ResNet \cite{he2016deep}, VGG-19 \cite{simonyan2014very}, and Inception \cite{szegedy2015going}. 
All these networks are pre-trained on the ImageNet dataset and fine-tuned to learn the generic matching function. 
Tao \textit{et al.} analysed VGG-16 and AlexNet models in SINT \cite{SINT16}.
Results show performance gap in tracking between these two pre-trained networks. 
It is also observed that direct replacement of powerful deep architectures does not bring performance improvement in the classical SNs-based tracking methods \cite{SIAMRPN19}.

To address this issue, Li \textit{et al.} investigated the main reason behind this issue and proposed a ResNet-driven SiamRPN++ tracker \cite{SIAMRPN19}.
In SNs, when modified AlexNet is employed without zero-padding \cite{SIAMFC16}, the learnt spatial feature representations of the target does not satisfy the spatial translational invariance constraint.
Therefore, an effective sampling technique is developed to fulfill this spatial invariance constraint. 
Leveraging the powerful deep ResNet architecture, the performance of many siamese trackers are improved. 
Zhang \textit{et al.} also investigated the same problem and proposed SiamDW in which a shallow backbone AlexNet is replaced with deep networks including Inception, VGG-19, and ResNet \cite{SIAMDW19}.
It is investigated that apart from features padding, receptive fields of neurons and network strides are also main reasons why such a deeper network cannot directly replace shallow networks.
The results presented in both studies demonstrated an excellent performance compared to classical SNs-based trackers.
With these foundations, recent trackers including SiamCAR \cite{SIAMCAR20}, Ocean \cite{OCEAN20}, and SiamBAN \cite{SIAMBAN20} etc., have also employed powerful deep architectures. 
\textit{The ResNet backbone has become the established and preferred alternative for Siamese tracking, thanks to its simplicity and strong performance. However, recent advances in vision transformer networks \cite{dosovitskiy2020image,liu2021swin} are expected to have a substantial impact in the tracking community in the coming years.}

\subsubsection{\textbf{Offline Training}}
\label{sec:training}
As discussed above, training data is crucial for VOT therefore it is very difficult to learn a robust matching function in SNs \eqref{eqn20}.

To handle this issue, the tracking community has made outstanding progress by leveraging images and video datasets to learn the generic relationships between the objects. 
Object detection, image classification, and object segmentation datasets including ImageNet ILSVRC2014 \cite{russakovsky2014imagenet}, ILSVRC2015 \cite{russakovsky2015imagenet}, COCO \cite{lin2014microsoft}, YouTube-BB \cite{real2017youtube},  and YouTube-VOS \cite{xu2018youtube}, have been widely used in SNs. 
These datasets sufficiently cover a good amount of semantics and do not focus on particular objects, otherwise, the tuned network parameters will overfit to particular object categories in siamese training. 
The datasets are typically annotated with bounding boxes of a target object in every frame.
Recently, the tracking community has compiled more diverse small and large-scale tracking benchmark datasets containing tens of thousands of annotated bounding boxes. 
These include the LaSOT \cite{fan2019lasot}, GOT-10K  \cite{huang2019got}, and TrackingNet  \cite{muller2018trackingnet} which are also used for offline training by recent trackers \cite{SIAMBAN20, SIAMRCNN20, CSA20}.
Interested readers can find more details about different training datasets utilized by different Siamese trackers in Table I of the supplementary material.

Unlike DCF paradigm, the standard Siamese formulation cannot exploit the appearance of known distractor objects during tracking. 
Siamese approaches therefore often struggle when objects similar to the target itself present. 
This occurs when, for instance, other objects of the same semantic class are in the view.
Early Siamese trackers (SiamFC \cite{SIAMFC16} and SiamRPN \cite{SIAMRPN18} particularly) only sample pairs of training images from the same video during training. This sampling strategy does not focus on challenging cases, with sementically similar distractor objects.
To address this issue, hard negative mining techniques have been developed in the literature. 
For instance, Zhu \textit{et al.} introduced hard negative mining technique in DaSiamRPN \cite{DASIAMRPN18} to overcome data imbalance issue by including more semantic negative pairs into the training process. 
The constructed negative pairs consist of labelled targets both in the same and different categories. 
This technique assisted DaSiamRPN to overcome drifting by focusing more on fine grained representations.
With the same spirit, Voigtlaender \textit{et al.} proposed another hard negative mining technique using an embedding network and a nearest neighbor approximation \cite{SIAMRCNN20}. 
For every ground truth target bounding box, the embedding vector is extracted for a similar target appearance using pre-trained network. 
The indexing structure is then employed to estimate the approximate nearest neighbors and use them to estimate the nearest neighbors of the target object in the embedding space.

\textit{The recent trend of utilizing more training data and designing data mining techniques have demonstrated excellent tracking performance on multiple benchmarks, opening many doors to explore more sophisticated techniques in Siamese training.}

\subsubsection{\textbf{Online Model Update}}
\label{sec:online}
In SiamFC \cite{SIAMFC16}, the target template is initialized in the first frame and then kept fixed during the remainder of the video. 
The tracker does not perform any model update and therefore, the performance totally relies on the general matching ability of the SN.
However, appearance changes in the presence of tracking challenges are often large and failing to update the model leads to failure of the tracker. 
In such scenarios, it is important to adapt the model to the current target appearance.
In the literature, the tracking community has also proposed potential solutions in this direction 

\noindent \textbf{Moving Average Update Method:} Many recent SOTA trackers including GOTURN \cite{GOTURN16}, SINT \cite{SINT16} and SiamAttn \cite{SIAMATTN20} etc., employ a simple linear update strategy using a running average with a fixed learning rate. 
While it provides a simple means of integrating new information, the trackers cannot recover from drift due to constant update rate and simple linear combination of previous appearance templates.

\noindent \textbf{Learning Dynamic SN Method:} Guo \textit{et al.} proposed the DSiam tracker and designed dynamic transformation matrices \cite{DSIAM17}. 
Two distinct online transformation matrices including target appearance variation and background suppression are incorporated within the classical SN.
Both matrices are solved in the Fourier domain with a closed-form solutions.
DSiam provides effective online learning however it ignores the historical target variations which is important for a smoother adaptation of the exemplar template.

\noindent \textbf{Dynamic Memory Network Method:} Yang \textit{et al.} proposed the MemTrack that dynamically writes and reads previous templates to cope with target appearance variations \cite{MEMTRACK18}.
A long term short term memory is used as a memory controller. 
The input to this network is the search feature map and the network outputs the control signals for the reading and writing process of the memory block.
An attention mechanism is also applied with a gated residual template to control the amount of retrieved memory that is used to combine with the initial template.
This method enables the tracker to memorize long-term target appearance.
However, it only focuses on combining the previous target features, ignoring the discriminative information in background clutter, which leads to an accuracy gap in the presence of drastic target variations. 

\noindent \textbf{Gradient-Guided Method:}
Li \textit{et al.} proposed GradNet in which gradient information is encoded for updating the target template via feedforward and backward operations \cite{GRADNET19}. 
The tracker utilizes the information from the gradient to update the template in the current frame and then incorporates the adaptation process to simplify the process of gradient-based optimization. 
Unlike aforementioned methods, this method makes full use of the discriminative information in backward gradients instead of just integrating previous templates. 
This results in a performance improvement as compared to other methods however, computing gradient in a backpropagation manner introduces a computational burden.

\noindent \textbf{UpdateNet Method:} Zhang \textit{et al.}, proposed the UpdateNet \cite{UPDATENET19} which learns a generic function $\theta$ according to $\Tilde{z}_{t}=\theta ( \tilde{z}_{gt}, \tilde{z}_{t-1}, z_{t})$
The learned function $\theta$ computes the updated template based on initial ground-truth template $z_{gt}$, the last accumulated template $\tilde{z}_{t-1}$ and the template $z_{t}$ extracted from the predicted target location in the current frame.
In general, the function updates the previous accumulated template $\tilde{z}_{t-1}$ by integrating the new information given by the current frame $z_{t}$.
Therefore, $\theta$ can be adapted to the specific updating requirements of the current frame, based on the difference between the current and accumulated templates.
Moreover, it also considers the initial template  $z_{gt}$ in every frame, which provides highly reliable information and increases robustness against model drift. 
The function $\theta$ is implemented as a CNN, which grants great expressive power and the ability to learn from large amounts of data. 
The results demonstrate excellent performance compared to SiamFC \cite{SIAMFC16} and DaSiamRPN \cite{DASIAMRPN18} and it has also recently been adopted by CSA tracker \cite{CSA20}.

\textit{While a number of techniques for model update have been proposed, simply using no update has remain a surprisingly robust and popular alternative \cite{SIAMRCNN20,SIAMRPN19}.
Further research in this direction is required to develop simple, general, and end-to-end trainable techniques, which could further improve the robustness of Siamese tracking.}

\subsubsection{\textbf{Loss Functions}} 
\label{sec:loss}
The tracking performance also relies on the loss functions employed within the SNs. 
Different loss functions have been used in the SNs either for regression, classification or for both tasks.  
Below, we summarize these developments in more detail.

\noindent \textbf{Logistic Loss:} The classical SiamFC employed logistic loss defined in Eqs.~\eqref{eqn21}-\eqref{eqn23} \cite{SIAMFC16}. 
A variety of other trackers including DSiam \cite{DSIAM17}, RASNET \cite{RASNET18}, SA-SIAM \cite{SASIAM18}, CFNET \cite{CFNET17}, SiamDW \cite{SIAMDW19}, and GradNet \cite{GRADNET19} etc., have used logistic loss to train their models built upon SiamFC. 
This training method utilizes the pairwise relationship on image pairs by maximizing the similarity scores on target-positive pairs and minimizing them on target-negative pairs.

\noindent \textbf{Contrastive Loss:} The margin contrastive loss is defined as \cite{chopra2005learning}:
\begin{equation}
\begin{split}
\mathcal{L} (x_{i},z_{i},y_{xz})=\frac{1}{2}D^{2}+\frac{1}{2}(1-y_{xz})\max(0,\epsilon-D^{2}), \\
D= ||f_{\rho}(x_{i})-f_{\rho}(z_{i})||_{2},
\end{split}
\label{eqn26}
\end{equation}
where $\epsilon$ is the minimum distance margin that pairs depicting different objects should satisfy, $D$ is the Euclidean distance  of $l_{2}$-normalized feature representations, and $y_{xz}\in \{0, 1\}$ indicates whether $x_{i}$ and $z_{i}$ are the same object or not. 
The SINT tracker \cite{SINT16} employed the contrative loss while GOTURN \cite{GOTURN16} employed $L_{1}$ loss between the predicted and the ground-truth bounding boxes.

\noindent \textbf{Triplet Loss:} The above loss exploits the pairwise relationship between images only and ignores the underlying structural connections between the positive and negative instances of the target. 
Yan \textit{et al.} proposed SPLT tracker \cite{SPLT19} in which the triplet loss \cite{weinberger2009distance, schroff2015facenet} is employed during training.
The triplet loss is defined as
\begin{equation}
\begin{split}
\mathcal{L} (x,x^{p}_{i},x^{n}_{i})=\sum_{i}^{M} \Big[||f_{\rho}(x_{i})-f_{\rho}(x^{p}_{i})||_{2}^{2}- \\
||f_{\rho}(x_{i})-f_{\rho}(x^{n}_{i})||_{2}^{2}+\epsilon \Big]_{+},
\end{split}
\label{eqn27}
\end{equation}
where $x^{p}_{i}$ denotes the positive patch of the target image $x$ i.e., one of other images of the target and  $x^{n}_{i}$ is a negative patch of any other target or background.
$\mathcal{T}$ is the set of all possible triplets in the training set and has cardinality $M$.
The triplet loss not only can further mine the potential relationship among target, positive and negative instances, but also contains more robust similarity structure.
Dong \textit{et al.,} proposed SiamFC-Tri in which the probabilistic triplet loss is used \cite{SIAMFCTRI18}.
The Siam R-CNN tracker has also been trained using the same loss \cite{SIAMRCNN20}. 

\noindent \textbf{Cross Entropy Loss:} The classification component in the SNs are normally borrowed from object detection methods \cite{girshick2015fast}. 
To incorporate this branch, a cross-entropy loss ($\mathcal{L}_{cls}$) is used which is defined as:
\begin{equation}
\begin{split}
\mathcal{L}_{cls}=-\sum_{j}p_{o}^{*}\log(p_{o})+(1-p_{o}^{*})\log(1-p_{o}),
\end{split}
\label{eqn28}
\end{equation}
where $p_{o}$ is a predicted label and $p_{o}^{*}$ denotes the groundtruth label.
Li \textit{et al.,} proposed the SiamRPN tracker in which cross entropy loss is employed \cite{SIAMRPN18}.
Other trackers such as SiamRPN++ \cite{SIAMRPN19}, SiamAttn \cite{SIAMATTN20}, Ocean \cite{OCEAN20}, CLNET \cite{CLNET20}, SPM \cite{SPM19}, C-RPN \cite{CRPN19} etc., have also been built upon the SiamRPN tracker by training classification branch using the cross entropy loss.

\noindent \textbf{Regression Loss:} To train a regression network, three types of loss functions are employed including the smooth $L_{1}$ norm \cite{girshick2015fast}, the Intersection over Union (IoU) loss \cite{yu2016unitbox}, and regularized linear regression \cite{scholkopf2002learning}.
The smooth $L_{1}$ loss is used in Faster R-CNN for bounding box regression \cite{girshick2015fast}.
In the SiamRPN tracker \cite{SIAMRPN18}, this norm is used to train the regression branch.
Following this study, other trackers including SiamRPN++ \cite{SIAMRPN19}, SiamAttn \cite{SIAMATTN20}, CLNET \cite{CLNET20}, SPM \cite{SPM19}, and C-RPN \cite{CRPN19} have also trained the regression branch of the tracker using the $smooth_{L1}$ loss.
The IoU loss for regression $\mathcal{L}_{reg}^{IoU}$ is defined as \cite{yu2016unitbox}: 
\begin{equation}
\mathcal{L}_{reg}^{IoU}=-\sum_{j}\ln(\text{IoU}(p,r)),
\label{eqn31}
\end{equation}
where $p$ and $r$ denote the predicted and groundtruth bounding box corrdinates.
Chen \textit{et al.} proposed the SiamBAN tracker in which the regression branch of the network is trained using $\mathcal{L}_{reg}^{IoU}$ \cite{SIAMBAN20}.
The Ocean \cite{OCEAN20} and SiaMFC++ \cite{SIAMFC20} trackers have also utilized this loss during training.

\noindent \textbf{Multi-Task Loss:} For the joint training of classification and regression branches, multi-task loss has also been used in SNs.
For instance, a sum of the cross-entropy loss \eqref{eqn28} and the regression loss \eqref{eqn31} (i.e., $\mathcal{L}_{cls}+\mathcal{L}_{reg}$) is used in SiamRPN \cite{SIAMRPN18}. 
Trackers such as SiamRPN++ \cite{SIAMRPN19}, SiamAttn \cite{SIAMATTN20}, CLNET \cite{CLNET20}, and SPM \cite{SPM19} have also employed the multi-task training method. 
Moreover, a sum of cross entropy loss \eqref{eqn28} and IoU loss for regression \eqref{eqn31} (i.e., $\mathcal{L}_{cls}+\mathcal{L}^{IoU}_{reg}$) has also been utilized by Ocean \cite{OCEAN20}, SiamBAN \cite{SIAMBAN20}, and SiamFC++ \cite{SIAMFC20}. 

\noindent \textbf{Regularized Linear Regression:} To regularize SNs with a correlation filter as a separate layer, a linear regression loss defined in Eqs. \eqref{eqn9}-\eqref{eqn10} is employed in many Siamese trackers including CFNET \cite{CFNET17}, TADT \cite{TADT19}, RTINET \cite{RTINET18}, DSiam \cite{DSIAM17}, FlowTrack \cite{FLOWTRACK18}, UDT \cite{UDT19}, and UDT++ \cite{UDT20} etc. 
The ridge regression problem is then solved by a closed-form solution and the filter is trained in an end-to-end fashion.

\textit{Currently, there is no general consensus in the literature regarding the employed loss function. Instead, recent SOTA methods adopt different alternatives. Among the aforementioned approaches, the Cross Entropy loss have remained a popular choice, also for recent trackers \cite{SIAMRPN19}.}

\subsubsection{\textbf{Target State Estimation}}
\label{sec:targetscaleestimation}
Similar to DCFs-based trackers, the SNs also suffer from severe scale variations challenges.
The similarity function only learns the deep structural generic relationship between the images and it does not take into account the problem of scale changes. 
The tracking community has also shown very good progress in this direction and proposed potential solutions to handle it. 
In the subsections below, we discuss the proposed methods.

\noindent \textbf{Multiple Resolution Scale Search Method:} Similar to the DCFs-based trackers, this method has also been employed in many siamese trackers to cope with scale variation challenges. 
In classical SiamFC, multiple scales are searched in a single forward-pass by assembling a mini-batch of scaled images and then the maximum response is computed. 
Early SN-based trackers including RASNET \cite{RASNET18}, SA-Siam \cite{SASIAM18}, StructSiam \cite{STRUCTSIAM18}, UDT \cite{UDT19}, UDT++ \cite{UDT20}, TADT \cite{TADT19}, GradeNet \cite{GRADNET19}, RTINET \cite{RTINET18}, and FlowTrack \cite{FLOWTRACK18} have employed this method for scale estimation.

\noindent \textbf{Deep Anchor-based Bounding Box Regression Method:} To estimate the target bounding boxes more accurately, the object detection capabilities are introduced within the SN \cite{SIAMFC16}. 
Anchor-based bounding box regression using a Regional Proposal Network (RPN) \cite{girshick2015fast} efficiently predicts region proposals with a wide variety of scales and aspect ratios. 
It takes an input image and estimates a set of rectangular object proposals, each with an objectness score.
To do so, a small network is slided over the convolutional feature map output by the last convolutional layer. 
At each sliding-window location, the multiple region proposals are simultaneously predicted.
RPN is a fully convolutional network that simultaneously predicts object bounds and objectness scores at each position. 
The RPN is trained in an end-to-end manner to generate high-quality region proposals \cite{girshick2015fast}.

Li \textit{et al.} proposed the SiamRPN tracker \cite{SIAMRPN18} which incorporates an RPN component. 
The outputs of SiamRPN include one classification ($\mathcal{L}_{cls}$) and one regression ($\mathcal{L}_{reg}$) branches to regress the target bounding box for both position and scale estimation.
Results demonstrated superior tracking performance compared to classical trackers in the presence of RPN.
Many recent trackers such as DaSiamRPN \cite{DASIAMRPN18}, SiamRPN++ \cite{SIAMRPN19}, SiamDW \cite{SIAMDW19}, SPLT \cite{SPLT19}, C-RPN \cite{CRPN19}, SiamAttn \cite{SIAMATTN20}, CSA \cite{CSA20}, and SPM \cite{SPM19}, etc., are also built upon the same notion.

\noindent \textbf{Deep Anchor-free Bounding Box Regression Method:} Anchor-based bounding box regression assists Siamese trackers \cite{SIAMRPN18, SIAMRPN19} to handle changes in scale and aspect ratio and shows encouraging results. 
However, it needs to carefully design anchor boxes based on heuristic knowledge, which introduces many hyper-parameters and computational complexity. 
Therefore, it is essential to carefully design and fix the parameters of the anchor boxes. 

Anchor-free regression has also been proposed in object detection, which avoids hyper-parameters associated with the anchor boxes and is more flexible and general \cite{law2018cornernet, zhu2019feature}.
Anchor-free detectors directly find objects without preset anchors using keypoint-based methods \cite{law2018cornernet} and center-based methods \cite{zhu2019feature}.
The keypoint-based methods first locate the pre-defined keypoints and then perform bounding box regression on the objects. 
In center-based methods, the four distances from positives to the object boundary are predicted using the center on the region of object. 
The anchor-free detectors are able to eliminate those hyperparameters related to anchors and have achieved similar performance with anchor-based detectors, making them more powerful in terms of generalization ability.

Chen \textit{et al.} proposed the SiamBAN tracker \cite{SIAMBAN20} in which anchor-free box regression is employed to estimate the target scale. 
The tracker avoids hyper-parameters associated with target bounding boxes without any preset anchor boxes.
The tracker exploits the expressive power of the fully convolutional network to classify objects and regresses their bounding boxes in a unified manner.
Similar to SiamRPN \cite{SIAMRPN19}, SiamBAN includes a classification module which performs foreground-background classification on each point of the correlation layer and a regression module performs bounding box prediction on the corresponding position \cite{SIAMBAN20}.
Ocean \cite{OCEAN20} and SiamCAR \cite{SIAMCAR20} trackers have also utilized the same method for scale estimation.
\textit{The object detection capabilities have demonstrated outstanding progress in the target state estimation component. 
The recent trends of using RPNs and anchor-free bounding box regression have revealed to further explore these techniques in an end-to-end paradigm.} Fig. \ref{fig2_infl} shows  influential Siamese trackers in literature.

\subsection{Evolution of Siamese Networks to Segmentation-based Tracker}
The Siamese trackers also face severe challenges in the presence of deformable objects such as a person with spread out hands, rotated or axis aligned bounding boxes \cite{SIAMMASK19}.
These tracking challenges can be addressed by assisting trackers using the most accurate and well-defined target location models. 
Siamese trackers are quite fast and provide real-time performance while video segmentation approaches are slow and not real-times therefore, combining these two problems provide efficient solution for both tracking and segmentation.

Recently, SNs have also been extended to perform both video object segmentation and tracking \cite{SIAMMASK19}.
Wang \textit{et al.} present a SN to simultaneously estimate binary mask, bounding box, and the corresponding background-foreground scores.
This multi-stage deep network lacks the opportunity to process the visual tracking and target segmentation jointly to increase robustness.
Lu \textit{et al.}, employed unsupervised video object segmentation task in which a novel architecture is proposed based on the co-attention mechanism within the SN \cite{lu2019see}. 

\begin{figure*}[t!]
\centering
\includegraphics[width=\linewidth]{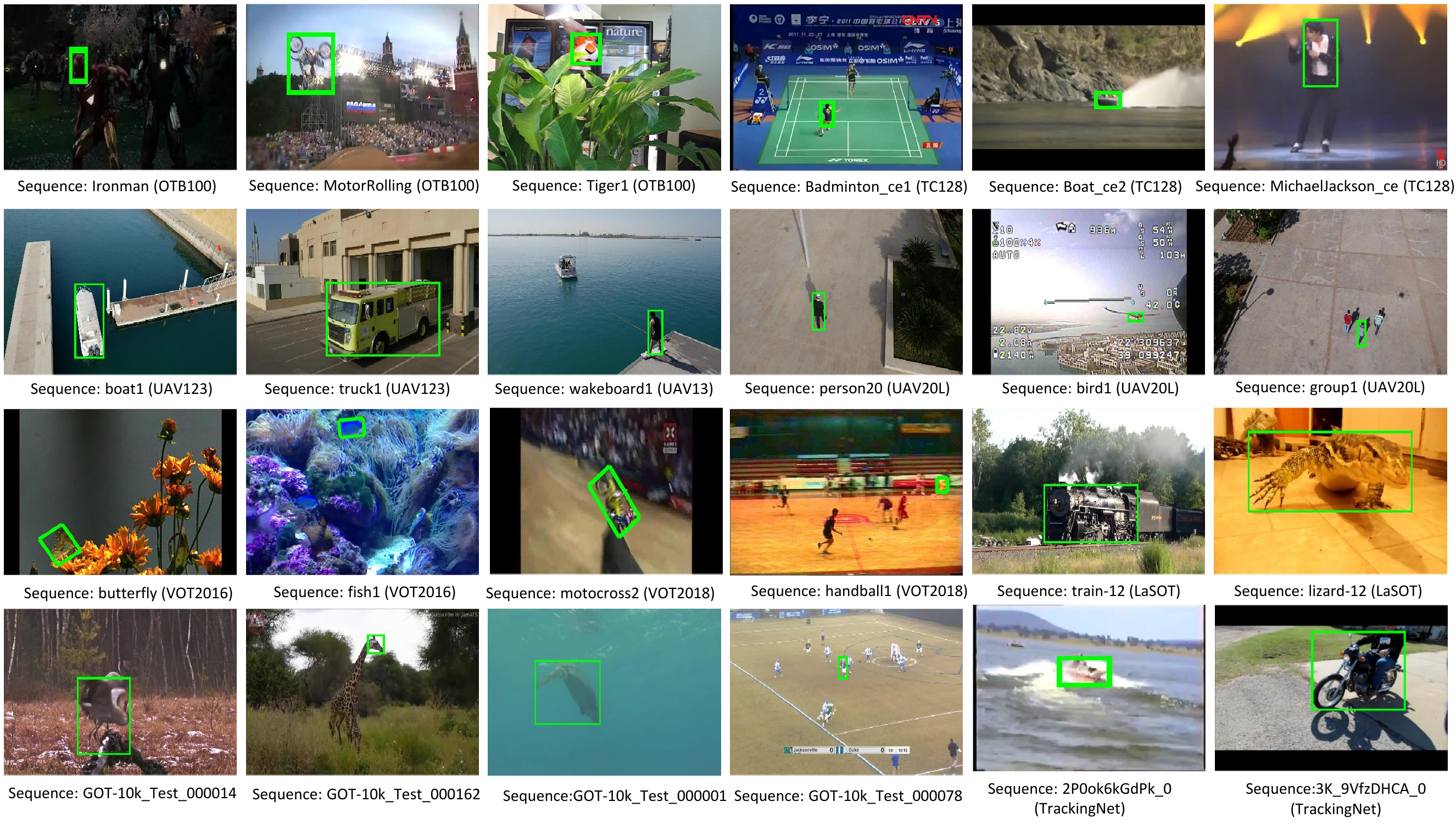}
\vspace{-1.7\baselineskip}
\caption{Sample images from the nine tracking benchmark datasets. The ground-truth bounding box annotation is overlaid. }
\label{fig2}
\vspace{-0.8\baselineskip}
\end{figure*}

\section{Experimental Comparison}
\label{sec:mainresults}
Here, we thoroughly analyze the performance of 59 DCFs and 33 Siamese-based trackers.
The performance of these trackers have been quantitatively compared on nine tracking benchmarks: Online Tracking Benchmark 100 (OTB100) \cite{wu2015object}, Temple Color 128 (TC128) \cite{liang2015encoding}, Unmanned Aerial Vehicle 123 (UAV) \cite{mueller2016benchmark}, Visual Object Tracking 2014 (VOT) \cite{VOT2014}, VOT2016 \cite{hadfield2016visual}, VOT2018 \cite{kristan2018sixth}, TackingNet \cite{muller2018trackingnet}, Large-scale Single Object Tracking (LaSOT) \cite{fan2019lasot}, and Generic Object Tracking 10,000 (GOT-10K) \cite{huang2019got}. 
Fig.~\ref{fig2} shows example frames from different tracking benchmarks.
The quantitative results of the compared trackers are either taken directly from respective papers or from other papers. 

\begin{table*}[t!]
\caption{Details of the nine benchmarks used in our experimental comparison.}
\vspace{-1.0\baselineskip}
\begin{center}
\makebox[\linewidth]{
\scalebox{0.47}{
\begin{tabu}{c|ccccccccccc}
\tabucline[2pt]{-}
Description&OTB100&TC128&UAV123 &UAV20L&VOT2014&VOT2016&VOT2018& VOT2018-LT&LaSOT&GOT-10K&TrackingNet\\\tabucline[2.0pt]{-}
Publication&PAMI2015 \cite{wu2015object}&TIP2015 \cite{liang2015encoding}&ECCV2016 \cite{mueller2016benchmark}&ECCV2016 \cite{mueller2016benchmark}&ECCV-W2014 \cite{VOT2014}&ECCV-W2016 \cite{hadfield2016visual}&ICCV-W2017 \cite{kristan2018sixth}&ECCV-W2018 \cite{kristan2018sixth}&CVPR2019 \cite{fan2019lasot}&PAMI2019 \cite{huang2019got}&ECCV2018 \cite{muller2018trackingnet}\\\tabucline[0.5pt]{-}
\multirow{3}{*}{Sequences}&&&&&&&&&Total: 1400&Total: 10k&Total: 30.643k\\&100&128&123&20&25&60&60&35&Training: 1120&Training: 9.34k&Training: 30.132k\\&&&&&&&&&Testing: 280&Testing: 420&Testing: 511\\\tabucline[0.5pt]{-}
\multirow{3}{*}{Boxes}&&&&&&&&&Total: 3.52M&Total: 1.5M&Total: 14M\\&58.61k&55.652k&113.476k&58.67k&10k&21.455k&21.356k&146.847k&Training: 2.8M&Training: 1.4M&Training: 14M\\&&&&&&&&&Testing: 685K&Testing: 56k&Testing: 226k\\\tabucline[0.5pt]{-}
Attributes&11&11&12&12&12&12&12&12&14&6&15\\\tabucline[0.5pt]{-}
\multirow{3}{*}{Classes}&&&&&&&&&&Total: 563&\\&22&27&9&9&11&24&24&21&70&Training: 480&21\\&&&&&&&&&&Testing: 84&\\\tabucline[0.5pt]{-}
Minimum Frames&71&71&109&1717&164&48&41&1389&1000&51&96\\\tabucline[0.5pt]{-}
Maximum Frames&3872&3872&3085&5527&1210&1507&1500&29700&11397&920&2368\\\tabucline[0.5pt]{-}
Average Resolution&$356 \times 530$&$461 \times 737$&$1231 \times 699$ &$1231 \times 699$&$448 \times 304$&$757 \times 480$&$758 \times 465$&$896 \times 468$&$632 \times 1089$&$929 \times 1638$&$591 \times 1013$\\\tabucline[0.5pt]{-}
Average Duration&19.68s&15.6s&30.48s&97.8s&13.68s&357.6s&355.9s&139.85s&83.57s&15s&16.7s\\\tabucline[0.5pt]{-}
\multirow{4}{*}{Attributes Name}&IV, SV, OCC, &IV, SV, OCC,&ARC, BC, CM,&ARC, BC, CM,&IV, SV, OCC,&OCO, SCO, AM,&OCO, SCO, AM,&IV, SV, OCC, &IV, SV, DEF, MB,&IV, SV,&IV, SV, DEF, MB,\\&DEF, MB, FM, &DEF, MB, FM,&FM, FOC, IV,&FM, FOC, IV,&DEF, MB, BC,&ARC, CM, MOC,&ARC, CM, MOC,&DEF, MB, BC, &FM, OV, BC, LR,&OCC, FM,&FM, IPR, OPR, OV, \\&IPR, OPR, OV, &IPR, OPR, OV, &LR, OV, POC,&LR, OV, POC,&ARC, CM, MOC, &DEF, MB, BC,&DEF, MB, BC,&ARC, CM, MOC,&ARC, CM, FOC,&SLO,&BC, LR, ARC, CM, \\&BC, LR.&BC, LR.&SOB, SV, VC.&SOB, SV, VC.&OCO, SCO, AM.&IV, SV, OCC.&IV, SV, OCC.&OCO, SCO, AM.&POC, VC, ROT.&ARC.&FOC, POC, SOB.\\\tabucline[0.5pt]{-}
Frame Rate&30 FPS&30 FPS&30 FPS&30 FPS&30 FPS&30 FPS&30 FPS&30 FPS&30 FPS&10 FPS&30 FPS\\\tabucline[0.5pt]{-}
\multirow{3}{*}{Overlapped Datasets}&OTB50&OTB50&&&OTB50&OTB100&OTB100, NUS-PRO \cite{nus_pro}&OTB100, NUS-PRO \cite{nus_pro}&YouTube&VOT&\\&VOT&VOT&VOT&VOT&TC128&TC128&ALOV++ \cite{smeulders2013visual}&ALOV++ \cite{smeulders2013visual}&ImageNet&WordNet \cite{miller1995wordnet}&YouTube-BB\\&TC128&OTB100&&&ALOV++ \cite{smeulders2013visual}&ALOV++ \cite{smeulders2013visual}&TC128, UAV123&TC128, UAV123&&ImageNet&\\\tabucline[2.0pt]{-}
\end{tabu}
}}
\end{center}
\label{table_dataset}
\vspace{-0.7\baselineskip}
\end{table*}

\subsection{Tracking Datasets}
To provide a standard and fair performance evaluation of object trackers, a number of benchmarks have been proposed with the passage of time. In addition to short-term tracking, several recent datasets provide both short and long-term tracking sequences.
The publicly available benchmark datasets contain a variety of tracking challenges, including Scale Variation (SV), Out-of-View (OV), DEFormation (DEF), Low Resolution (LR), Illumination Variation (IV), Out-of-Plane rotation (OPR), OCClusion (OCC), Background Clutter (BC), Fast Motion (FM), In-Plane Rotation (IPR), Motion Blur (MB), Partial OCclusion (POC), abrupt Camera Motion (CM), Aspect Ratio Change (ARC), Full OCclusion (FOC), Viewpoint Change (VC), Similar OBject (SOB), Object Color Change (OCC), Absolute Motion (AM), target ROTation (ROT), Scene COmplexity (SCO), Fast Camera Motion (FCM), Low Resolution Objects (LRO), and MOtion Change (MOC).
Table \ref{table_dataset} presents the description of each dataset employed in our experimental comparison. Next, we briefly describe each tracking dataset.

\subsubsection{\textbf{OTB100 Dataset}}
Wu \textit{et al.} proposed an object tracking benchmark known as OTB50 \cite{Wu2013CVPR}. 
This dataset consists of 51 video sequences with manually annotated bounding boxes in each frame. 
The sequences are categorized into 11 different tracking attributes (\ref{table_dataset}).
Later on, Wu \textit{et al.} extended OTB50 to OTB100 dataset by adding 49 more videos \cite{wu2015object}. 
OTB100 consists of 100 videos of 22 object categories with the same 11 tracking attributes as in OTB50. 
The average resolution in the OTB100 dataset is $356 \times 530$, while the video length ranges between 71 and 3872 frames.

\subsubsection{\textbf{TC128 Dataset}}
TC128 was introduced to evaluate the impact of color on visual tracking \cite{liang2015encoding}. 
It contains 128 fully annotated color video sequences of 27 object categories.
Out of 128 sequences, 78 sequences are different from OTB100, whereas the remaining 50 sequences are common in both datasets. 
TC128 also comprises 11 tracking attributes, similar to OTB100 (Table \ref{table_dataset}). 
The average resolution is $461 \times 737$, with 71 minimum and 3872 maximum number of frames.

\subsubsection{\textbf{UAV123 Dataset}}
UAV123 contains a realistic and synthetic HD video sequences captured by professional-grade UAV  \cite{mueller2016benchmark}. 
Videos are captured from low-altitude UAVs which are inherently different from sequences in other datasets. 
This dataset is divided into two subsets namely, UAV123 and UAV20L. 
The UAV123 contains 123 short sequences of 9 diverse object categories, with 109 minimum number of frames, and 3085 maximum number of frames. 
The UAV20L consists of 20 long videos of 5 object classes generated from a flight simulator.
These sequences contain 1717 minimum and 5527 maximum number of frames. 
Both dataset subsets contain an average resolution of $1231 \times 699$ and labelled with 12 attributes (Table \ref{table_dataset}).

\subsubsection{\textbf{VOT Dataset Series}}
The VOT dataset accompanies the annual VOT challenge competition to benchmark tracking performance \cite{kristan2016novel}.
Each frame in the VOT datasets is annotated with a rotated bounding box with a number of tracking challenges.
Here, we select VOT2016, VOT2018, and VOT2020 datasets as representatives of VOT series to compare the performance of different trackers.
A short description of VOT datasets is presented in Table \ref{table_dataset}.

\noindent \textbf{VOT2016 Dataset \cite{hadfield2016visual}:} This dataset contains 60 sequences, as in VOT2015 \cite{kristan2015visual}. 
The only difference is that the ground-truth bounding boxes in VOT2016 are more accurate than VOT2015, since a segmentation-based method is employed to generate more precise and accurate bounding boxes. Each sequence is per-frame annotated by different attributes, including OCC, IV, MOC, ARC, SCO, and FCM. 
The average resolution of sequences is $757 \times 480$, with 48 minimum and and 1507 maximum number of frames.

\noindent \textbf{VOT2018 Dataset:}  Different from VOT2016, this dataset consists of short and long-term challenge splits \cite{kristan2018sixth}. 
The VOT2018 Short-Term (VOT2018-ST) challenge consists of 60 sequences of 24 object categories, as in VOT2017 \cite{kristan2017visual}. 
The average resolution of sequences in short-term challenge is $758 \times 465$, with 41 minimum and 1500 maximum number of frames.
The long-term split consists of 35 long-term sequences.
The average resolution of sequences in long-term is $896 \times 468$, with 1389 minimum and 29700 maximum number of frames.

\noindent \textbf{VOT2020 Dataset:}  VOT2020 consists of five sub-sets, including sequestered short-term and long-term sequences \cite{kristan2020eighth}.
We use VOT2020 Short-term (VOT2020-ST) dataset to evaluate the performance of the trackers. The VOT2020-ST is the same as VOT2018-ST in terms of number of videos, classes, and attributes.
The only difference is that the target position is encoded by a segmentation mask with new performance evaluation measures and protocol.
The initial masks are obtained by a semi-automatic method and then all sequences are frame-by-frame manually corrected.
The main objective is to evaluate trackers for segmentation tasks that compliments features from both tracking and video object segmentation.

\subsubsection{\textbf{TackingNet Dataset}}
Muller \textit{et al.} proposed the TackingNet dataset for long-term tracking in the wild  \cite{muller2018trackingnet}.
It consists of 60,643 sequences with more than 14 million dense bounding box annotations. 
It covers 27 diverse object classes. 
The sequences are also represented by 15 tracking attributes.
The dataset is divided in to training, validation, and test splits.
The training split contains 30643 sequences, whereas test split consists of 511 videos. 
In test split, the average resolution of a sequence is $591 \times 1013$, with 96 minimum and 2368 maximum number of frames at 30fps.

\subsubsection{\textbf{LaSOT Dataset}}
Fan \textit{et al.} proposed the LaSOT dataset that consists of 1120 training sequences (2.8M frames) and 280 testing sequences (685K frames) \cite{fan2019lasot}.
All sequences are annotated with bounding boxes in every frame.
The object categories are selected from the ImageNet. 
It contains 70 diverse object categories and each category consists of 20 target sequences. 
The sequences are categorized according to 14 attributes, including ARC, BC, FCM, DEF, POC, ROT, and VC. 
The average resolution of a sequence is $632 \times 1089$. Moreover, the dataset contains very long sequences, ranging between 1000 and 11,397 number of frames.

\subsubsection{\textbf{GOT-10K Dataset}}
GOT-10K \cite{huang2019got} consists of 10,000 videos from semantic hierarchy of WordNet \cite{miller1995wordnet}. 
The aim is to provide a unified training and evaluation platform for the development of class-agnostic trackers with rich motion trajectories. 
The sequences are classified to 563 classes of moving objects, six tracking attributes, and 87 classes of motion to cover as many challenging patterns in real-world scenarios as possible.
GOT-10K is divided into training, validation, and test splits. 
The training split contains 9,340 sequences with 480 object categories, whereas test split consists of 420 videos with 83 object categories and each sequence with an average length of 127 frames. 
In test split, the average resolution of a sequence is $929 \times 1638$, with 51 minimum and 920 maximum number of frames at 10fps.

\subsection{Performance Evaluation Measures}
To compare the performance of trackers, different evaluation metrics have been proposed in the literature that evaluate the effectiveness in terms of robustness, accuracy and speed.

\noindent \textbf{Precision Plot:} The precision plot is based on the central location error which is defined as the average Euclidean distance between the predicted centers of the target object and the ground truth centers in a frame \cite{wu2015object}. 
However, this error does not compute the tracking performance accurately. Therefore distance precision is employed, which is defined as the percentage of frames where the target object is located within a center location error of $T$ pixels \cite{babenko2010robust}.
The trackers are ranked using this metric with a threshold of $T=20$ pixels.
The precision plot is generated by plotting the distance precision over a range of thresholds. 

\noindent  \textbf{Success Plot:} The precision metric only measures the localization performance of the tracker which is not accurate to measure the target scale variations \cite{wu2015object}. 
Instead of the center location error, the IoU is employed to measure the prediction error. Given the estimated bounding box $p$ and ground-truth bounding box $g$, the IoU is defined as  $\frac{p \cap g}{p \cup g}$. The success rate is thus the percentage if frames where the IoU is smaller than a $T$. 
The success plot is be generated by varying the overlap threshold from 0 to 1.
The trackers are ranked using area under the curve of the success plot.

\noindent \textbf{Normalized Precision Plot:} As distance precision is sensitive to the target scale, Muller \textit{et al.,} employed normalized precision metric to evaluate trackers based on the relative error \cite{muller2018trackingnet}.
It computes errors relative to the target size instead of considering the absolute distance i.e., $||W(p_{x},p_{y})-(g_{x},g_{y})||$, where $W=\textrm{diag}(g_{x},p_{y})$.
This relative error is then plotted in the range of 0 to 0.5. The area under this curve is called normalized precision which is used to rank trackers.

\noindent \textbf{Average Overlap:} This metric estimates the average of overlaps between the ground-truth and estimated bounding boxes, as in the success plot \cite{huang2019got}.

\noindent \textbf{SR$_{0.50}$ and SR$_{0.75}$:} These metrics denote the success rate that measures the percentage of successfully tracked frames, where the overlap precision exceeds a threshold of 0.50 and 0.75

The One pass evaluation criteria is used as defined in \cite{wu2015object} to measure the tracking performance in terms of precision and success plots on OTB100, TC128, UAV123, and LaSOT datasets.
The trackers on these datasets are evaluated by initializing bounding box on the first frame and letting it run until the end of the sequence.

In the VOT series, a tracker is reset once it drifts off the target. Following the VOT evaluation protocols \cite{VOT2014, hadfield2016visual, kristan2017visual}, the trackers are compared in terms of Accuracy (A), Robustness (R), and Expected Average Overlap (EAO) metrics. 
A is the average overlap between the predicted and ground truth bounding boxes during successful tracking periods.
R measures how many times the tracker loses the target (fails) during tracking. 
A reset mechanism starts after some frames once tracker losses the target object. 
EAO is an estimator of the average overlap a tracker is expected to attain on a large collection of short-term sequences with the same visual properties as the given dataset.

\def \widthfortwocolumns {0.49\columnwidth}
\def \heightforvspace {0.025in}
\begin{figure}[t]
\centering
\subfigure{\begin{minipage}{\widthfortwocolumns}
\includegraphics[width=\textwidth,height=\linewidth]{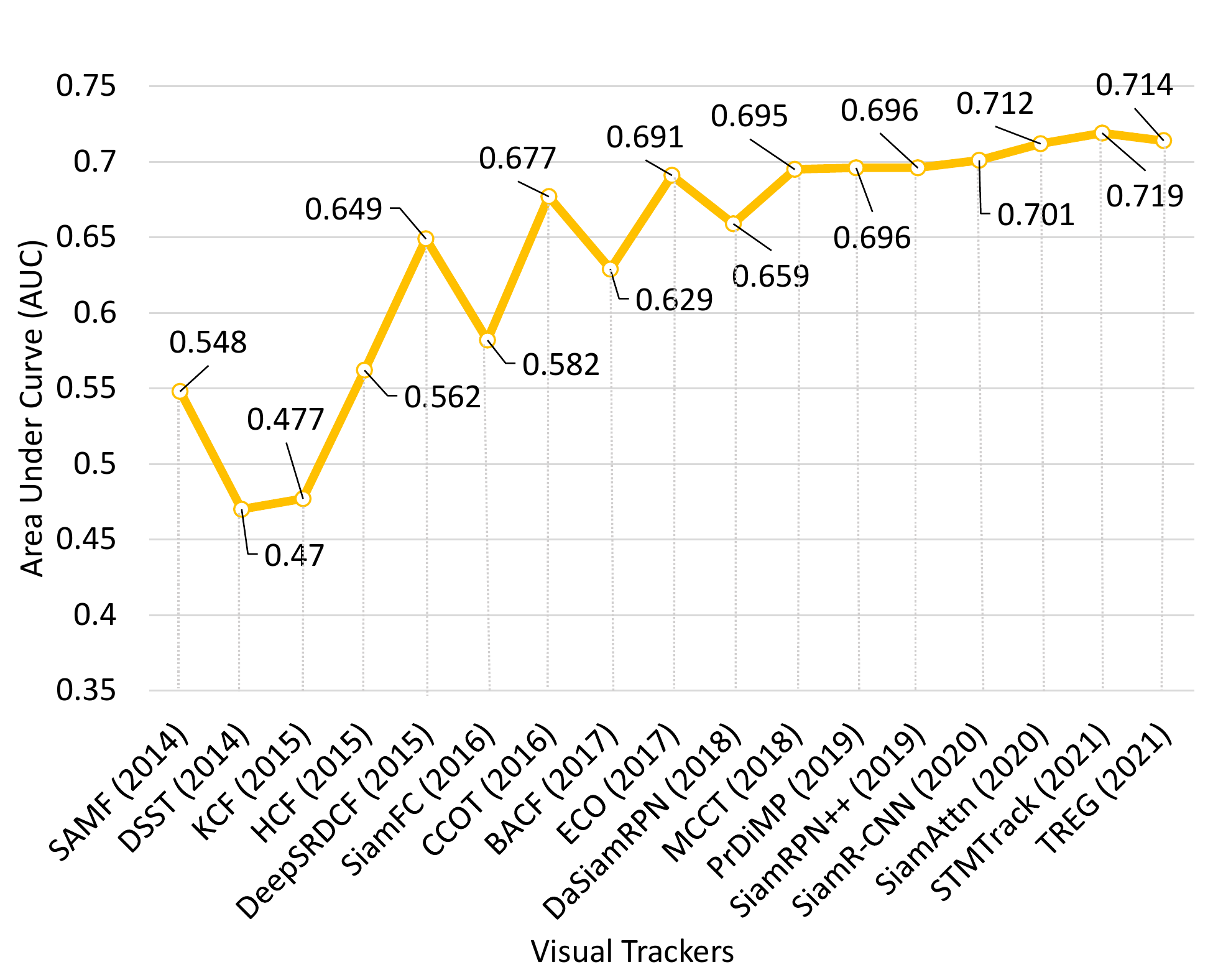}\vspace{-0.5\baselineskip}\caption*{(a) OTB100}\vspace{\heightforvspace}
\includegraphics[width=\textwidth,height=\linewidth]{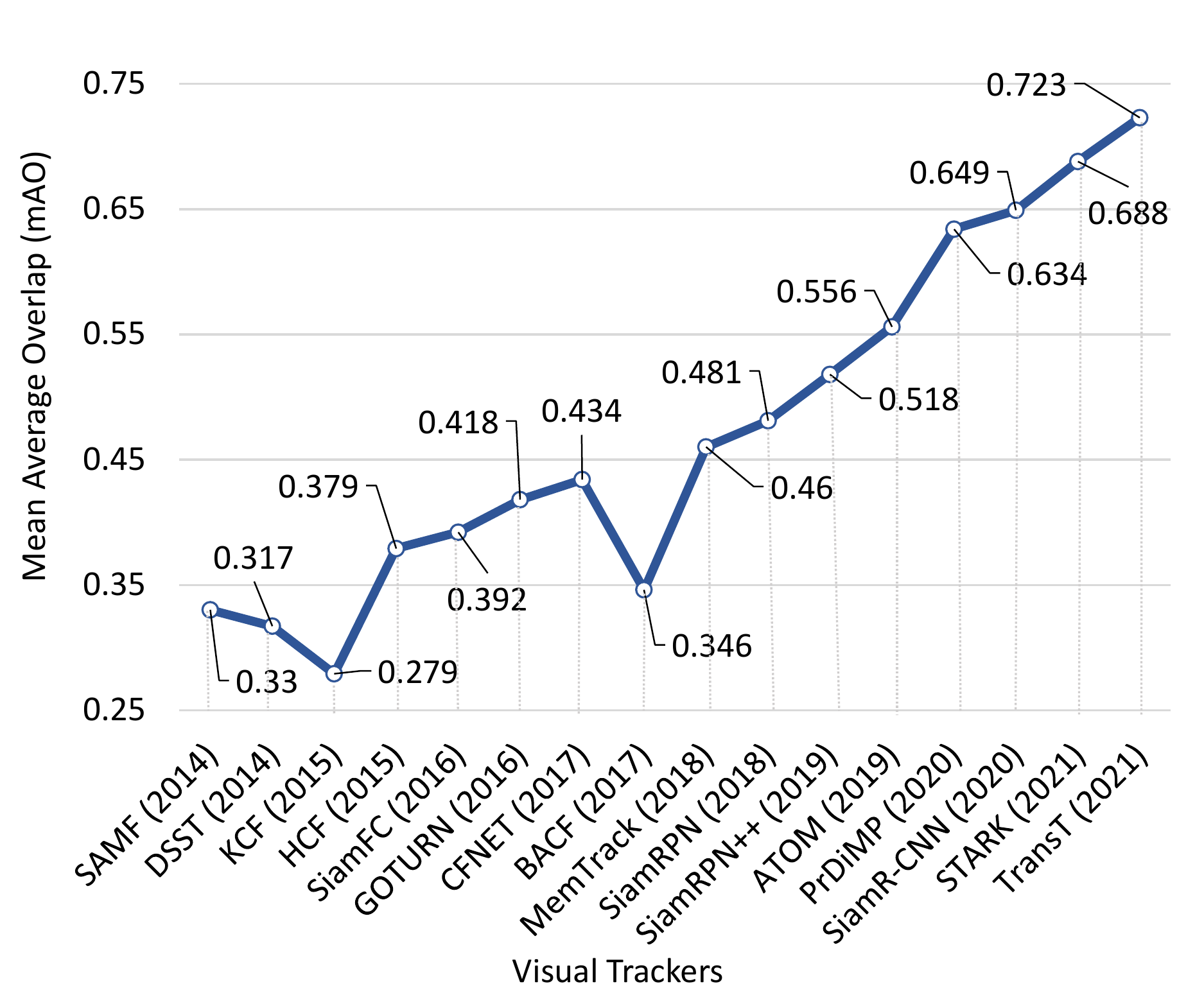}\vspace{-0.5\baselineskip}\caption*{(c) GOT-10K}\vspace{\heightforvspace}
\end{minipage}} 
\subfigure{\begin{minipage}{\widthfortwocolumns}
\includegraphics[width=\textwidth,height=\linewidth]{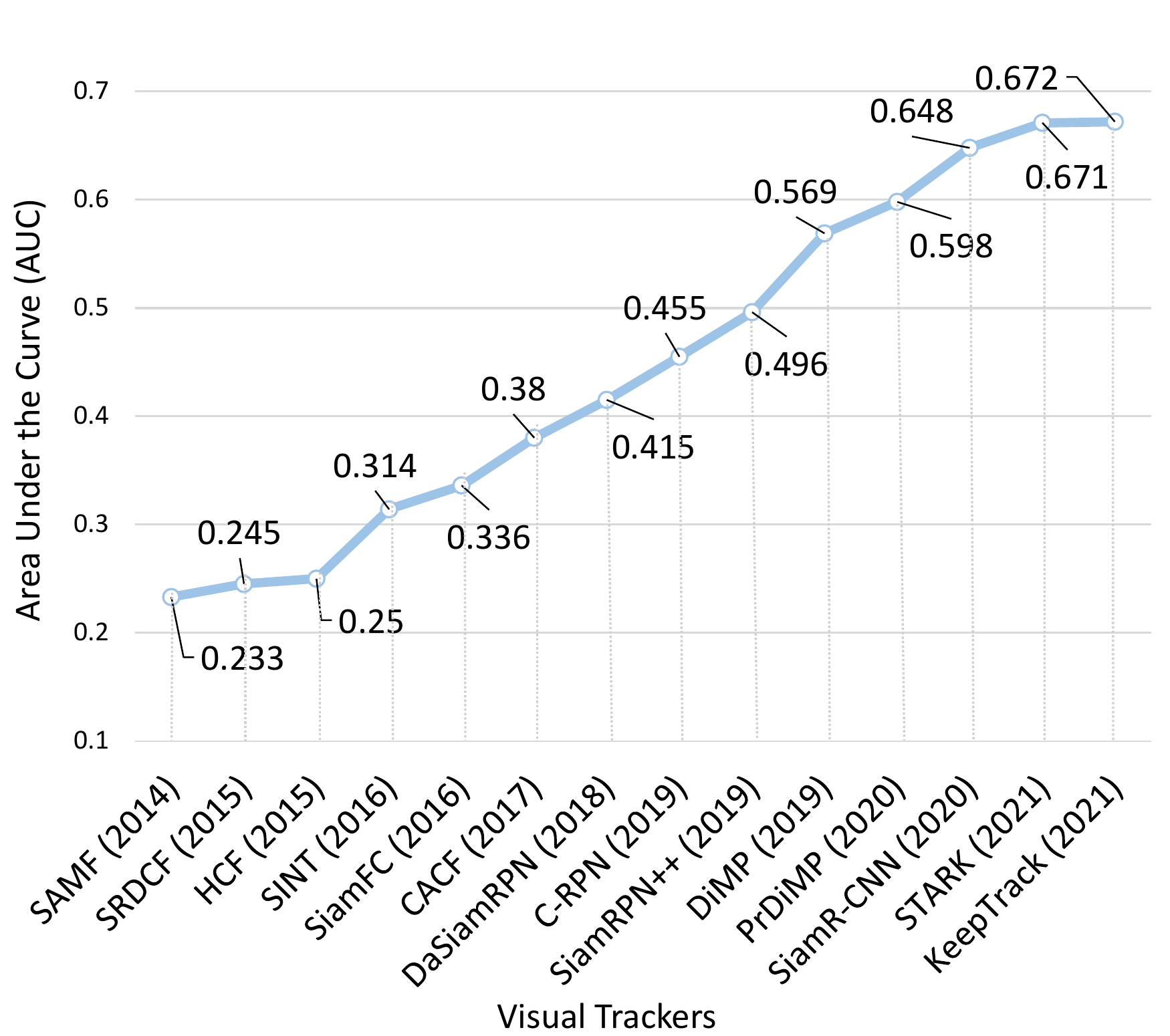}\vspace{-0.5\baselineskip}\caption*{(b) LaSOT}\vspace{\heightforvspace}
\includegraphics[width=\textwidth,height=\linewidth]{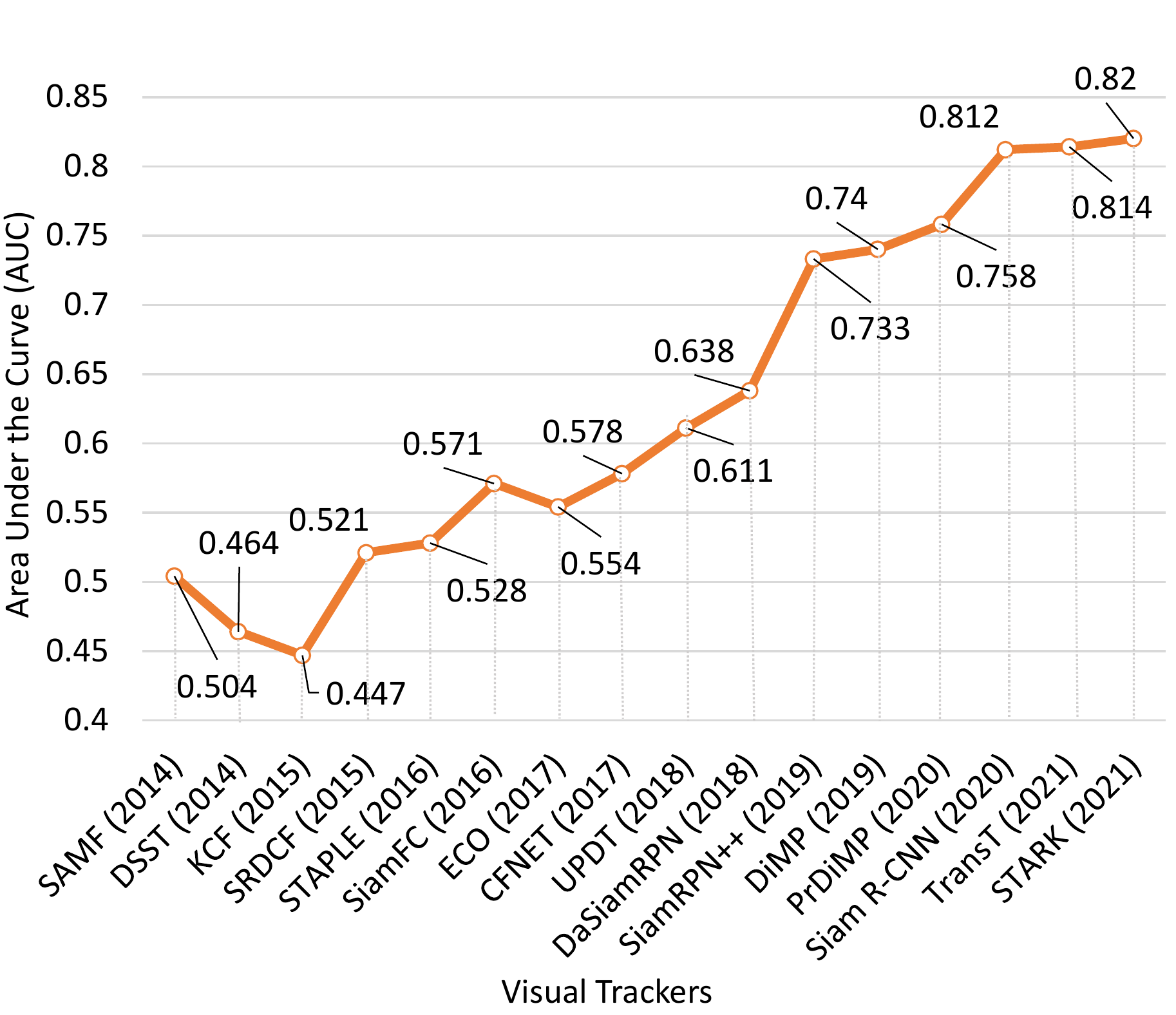}\vspace{-0.5\baselineskip}\caption*{(d) TrackingNet}\vspace{\heightforvspace}
\end{minipage}}  
\vspace{-1.4\baselineskip}
\caption{Tracking performance trends on popular benchmarks: (a) OTB100, (b) LaSOT, (c) GOT-10k and (d) TrackingNet, during 2014 - 2020.
The representative trackers are: SAMF \cite{SAMF14}, DSST \cite{DSST14}, KCF \cite{KCF15}, HCF \cite{HCF15}, SRDCF \cite{SRDCF15}, DeepSRDCF \cite{DEEPSRDCF15}, STAPLE \cite{STAPLE16}, SiamFC \cite{SIAMFC16}, GOTURN \cite{GOTURN16}, SINT \cite{SINT16}, CACF \cite{CACF17}, ECO \cite{ECO17},
CCOT \cite{CCOT16}, BACF \cite{BACF17}, CSR-DCF \cite{CSRDCF17}, CFNET \cite{CFNET17}, MemTrack \cite{MEMTRACK18}, UPDT \cite{UPDT18}, SiamRPN \cite{SIAMRPN18}, DaSiamRPN \cite{DASIAMRPN18}, DRT \cite{DRT18}, C-RPN \cite{CRPN19}, MCCT \cite{MCCT18}, DiMP \cite{DIMP19}, SiamMask \cite{SIAMMASK19}, D3S \cite{D3S20}, PrDiMP \cite{PRDIMP20}, ATOM \cite{ATOM19}, SiamRPN++ \cite{SIAMRPN19}, SiamR-CNN \cite{SIAMRCNN20}, and SiamAttn \cite{SIAMATTN20}.}
\label{fig_perftrend}
\vspace{-1.5\baselineskip}
\end{figure}

\subsection{Quantitative Comparison}
Tables \ref{table1_dcf}, \ref{table2_dcf}, \ref{table3_seg}, and \ref{table3_dcf} present the performance comparison of representative DCF-based trackers on nine tracking benchmarks.\footnote{For a more detailed comparison, please see supplementary material.\label{footnote:hw_regions}} 
While earlier DCF-based trackers employing deep features achieve promising performance on OTB100, they provide inferior results on recent more challenging large-scale datasets such as LaSOT. 
For instance, ECO achieves an impressive PR score of 91.0$\%$ on OTB but obtains a PR score of only 30.1$\%$ on LaSOT.
In contrast, recent end-to-end DCF frameworks such as, DiMP and its successor PrDiMP achieve impressive performance on OTB100 as well as on LaSOT. 
For instance, PrDiMP achieves PR scores of 90.3$\%$, 87.8$\%$ and 60.9$\%$ on OTB100, UAV123, and LaSOT, respectively. 
Among existing DCF-based trackers, DiMP and PrDiMP achieve superior results on most benchmarks. 
PrDiMP achieves top performance on UAV123, LaSOT, and GOT-10K, while also achieving competitive results (among top-three) on OTB100, VOT2016, and VOT2018-ST. 
The success of these modern DCF trackers (DiMP and \cite{PRDIMP20,KYS20}) is due to their efficient end-to-end trainable architectures that are capable of learning a discriminative target model prediction by fully utilizing both target and background appearance information. 
These trackers employ a dedicated optimization process to learn a powerful model in few iterations. 
For instance, PrDiMP utilizes a more general Newton approximation for addressing KL-divergence objective. 
Further, these modern trackers comprise a dedicated target estimation component to perform deep bounding box regression and also circumvent the problem of boundary artefacts.

Tables \ref{table1_dcf}, \ref{table2_dcf}, \ref{table3_seg}, and \ref{table3_dcf} also present the performance comparison of representative Siamese trackers on nine benchmarks \footref{footnote:hw_regions}. 
Among recent Siamese methods, we observe trackers to focus on different fundamental issues such as, online model update, re-detection components, improved region refinement, effective box regression, and bridging the gap between object tracking and object segmentation. 
For instance, SiamAttn introduces an attention mechanism to adaptively update the target template and obtains the best performance on OTB100, UAV123, VOT2016, while also achieving competitive results (among top-three) on LaSOT, VOT2018-ST, and TrackingNet. 
SiamAttn obtains AUC scores of 71.2$\%$, 65.0$\%$, 56.0$\%$, and 75.2$\%$ on OTB100, UAV123, LaSOT, and TrackingNet, respectively. 
Further, it obtains EAO scores of 53.7$\%$ and 47.0$\%$ on VOT2016 and VOT2018-ST, respectively. 
SiamR-CNN introduces a re-detection architecture combined with a tracklet-based dynamic programming scheme and obtains top performance on TC128 (64.9$\%$ AUC score), LaSOT (64.8$\%$ AUC score), GOT-10K (64.9$\%$ mAO score) and TrackingNet (81.2$\%$ AUC score), while also achieving competitive results (among top-three) on other datasets. 
Ocean introduces an approach to refine the imprecise bounding-box predictions along with learning object-aware features and obtains the top performance on VOT2018-ST (48.9$\%$ EAO score). 
D3S is a single-shot segmentation tracker employing two target models with complementary properties and obtains the best results on VOT2018-ST and VOT2020-ST (48.9$\%$ and 43.9$\%$ EAO score).

Figure \ref{fig_perftrend} shows the tracking performance improvement trend on different benchmarks (OTB100, LaSoT, GOT-10k, and TrackingNet) in recent years. 
We can observe that the performance on OTB100 has saturated in recent years with several visual trackers obtaining over 90$\%$ PR score (Table \ref{table1_dcf}), likely due to numerous relatively easy videos. 
While, the recently introduced LaSOT, GOT10K, and TrackingNet all show a similar trend with consistent improvements obtained by recent trackers on these datasets. We also observed a similar trend on the VOT dataset in Figure \ref{fig_VOT}. For instance, the best reported AUC score on LaSOT is still around 65$\%$. Similarly, there is still a significant room to further improve the tracking performance in the VOT dataset, despite witnessing an impressive leap in performance in recent years. This suggests that these new challenging benchmarks are still very challenging for SOTA trackers and their introduction is significantly contributing to pushing the boundaries of visual tracking research.

\textit{The aforementioned datasets also have radically different properties and characteristics.
LaSOT and UAV123 contain long sequences and multiple distractors. 
Trackers achieving high performance here demonstrate substantial robustness and re-detection capabilities.
We observe that the recent trackers DiMP and PrDiMP achieve strong results, and that the distractor-aware track generation in SiamR-CNN improves the robustness in such scenarios.
In contrast to LaSOT, TrackingNet, and GOT10k contain short sequences where robustness and re-detection capability is of much lesser importance.
Instead, these datasets reward trackers with highly accurate bounding box prediction, such as SiamR-CNN and PrDiMP. 
Among Siamese trackers, we obsserve that SiamR-CNN and SiamAttn achieve the most consistently good results across several datasets.
The exception being LaSOT in case of SiamAttn, while SiamR-CNN struggles on VOT. 
Among DCF-based methods, PrDiMP consistently achieves SOTA results across all evaluated datasets. }

\begin{table*}[t!]
\caption{Performance comparison of representative DCF and Siamese trackers, in terms of Precision Rate (PR) at a threshold of 20 pixels, Normalized Precision Rate (NPR), and Area Under the Curve (AUC), on OTB100, TC128, UAV123, UAV20L, and LaSOT datasets. 
For both DCF and Siamese trackers, the best two results are in red and blue font, respectively. }
\vspace{-1.7\baselineskip}
\begin{center}
\makebox[\linewidth]{
\scalebox{0.80}{
\begin{tabu}{|[2.0pt]c|[2.0pt]c|[2.0pt]c|c|[2.0pt]c|c|[2.0pt]c|c|[2.0pt]c|c|[2.0pt]c|c|c|[2.0pt]c|c|[2.0pt]}
\tabucline[2.5pt]{3-16}
\multicolumn{2}{c|[2.0pt]}{}&\multicolumn{2}{c|[2.0pt]}{OTB100}&\multicolumn{2}{c|[2.0pt]}{TC128}&\multicolumn{2}{c|[2.0pt]}{UAV123}&\multicolumn{2}{c|[2.0pt]}{UAV20L}&\multicolumn{3}{c|[2.0pt]}{LaSOT}&\multicolumn{2}{c|[2.0pt]}{Speed (fps)}\\\tabucline[2.5pt]{-}
\textbf{DCF Trackers}&\textbf{Publication}&\textbf{PR}&\textbf{AUC}&\textbf{PR}&\textbf{AUC}&\textbf{PR}&\textbf{AUC}&\textbf{PR}&\textbf{AUC}&\textbf{PR}&\textbf{NPR}&\textbf{AUC}&\textbf{CPU}&\textbf{GPU}\\\tabucline[2.5pt]{-}
MOSSE \cite{MOSSE10}&CVPR2010&0.414&0.311&-&-&0.466&0.297&0.339&0.223&-&-&-&\textcolor{red}{\textbf{669}}&-\\\tabucline[0.5pt]{-}
%CSK \cite{CSK12}&ECCV2012 &0.519&0.382&0.465&0.350&0.488&0.311&0.309&0.194&0.131&0.149&0.149&299&-\\\tabucline[0.5pt]{-}
%%STC \cite{STC14}&ECCV2014&0.507&0.319&-&-&-&-&-&-&0.137&0.156&0.138&\textcolor{blue}{\textbf{350}}&-\\\tabucline[0.5pt]{-}
SAMF \cite{SAMF14}&ECCV2014 &0.743&0.548&-&0.467&0.592&0.396&0.457&0.317&0.203&0.239&0.233&7&-\\\tabucline[0.5pt]{-}
%ACA \cite{ACA14}&CVPR2014 &0.595&0.440&0.505&0.366&-&-&-&-&0.163&0.177&0.170&105&-\\\tabucline[0.5pt]{-}
%DSST \cite{DSST14}&BMVC2014 &0.693&0.470&0.534&0.405&0.586&0.356&0.456&0.270&0.189&0.213&0.207&25.4&-\\\tabucline[0.5pt]{-}
KCF \cite{KCF15}&PAMI2015 &0.695&0.477&0.588&0.418&0.523&0.331&0.311&0.198&0.166&0.190&0.178&172&-\\\tabucline[0.5pt]{-}
%RPT \cite{RPT15} &CVPR2015&0.754&0.534&-&-&-&-&-&-&-&-&-&4.1&-\\\tabucline[0.5pt]{-}
%CFLB \cite{CFLB15}&CVPR2015 &-&0.415&-&0.347&-&-&-&-&-&-&-&100&-\\\tabucline[0.5pt]{-}
HCF \cite{HCF15}&ICCV2015 &0.837&0.562&0.703&0.488&-&-&0.500&0.299&0.241&0.286&0.250&-&10.4\\\tabucline[0.5pt]{-}
%MUSTer \cite{MUST15}&CVPR2015 &0.774&0.577&0.636&0.471&0.591&0.392&0.514&0.329&-&-&&4&-\\\tabucline[0.5pt]{-}
%SRDCF \cite{SRDCF15}&ICCV2015 &0.776&0.597&0.696&0.509&0.676&0.464&0.507&0.343&0.219&0.248&0.245&5&-\\\tabucline[0.5pt]{-}
DeepSRDCF \cite{DEEPSRDCF15} &ICCV-W2015&0.851&0.649&0.740&0.536&-&-&-&-&-&-&-&-&0.2\\\tabucline[0.5pt]{-}
%LCT \cite{LCT15}&CVPR2015&0.762&0.564&0.606&0.455&-&-&0.368&0.260&0.190&0.209&0.221&27.4&-\\\tabucline[0.5pt]{-}
fDSST \cite{fDSST16}&PAMI2016 &0.726&0.551&0.575&0.435&-&-&0.422&0.300&0.184&0.208&0.203&54.3&-\\\tabucline[0.5pt]{-}
%HDT \cite{HDT16}&CVPR2016 &0.835&0.564&0.686&0.452&-&-&-&-&-&-&-&-&10\\\tabucline[0.5pt]{-}
STAPLE \cite{STAPLE16}&CVPR2016 &0.784&0.581&0.665&0.498&-&-&-&-&0.239&0.278&0.243&80&-\\\tabucline[0.5pt]{-}
%CCOT \cite{CCOT16}&ECCV2016 &0.898&0.677&0.774&0.597&-&-&-&-&-&-&-&-&0.3\\\tabucline[0.5pt]{-}
%SCT \cite{SCT16}&CVPR2016 &0.768&0.534&-&-&-&-&-&-&0.179&-&0.191&37&-\\\tabucline[0.5pt]{-}
%CFAT \cite{CFAT16}&ECCV2016 &0.790&0.680&-&-&-&-&-&-&-&-&-&4.25&-\\\tabucline[0.5pt]{-}
%SRDCFdecon \cite{SRDCFDECON16}&CVPR2016 &0.825&0.634&0.729&0.541&-&-&-&-&-&-&-&3&-\\\tabucline[0.5pt]{-}
%BACF \cite{BACF17}&ICCV2017 &0.861&0.629&0.660&0.519&0.592&0.396&-&-&0.239&0.283&0.259&35.3&-\\\tabucline[0.5pt]{-}
%CACF \cite{CACF17}&CVPR2017 &0.810&0.598&-&-&-&-&-&-&0.235&-&0.380&35.2&-\\\tabucline[0.5pt]{-}
%CSR-DCF \cite{CSRDCF17}&CVPR2017 &0.733&0.587&0.715&0.507&-&-&-&-&0.220&0.254&0.244&13&-\\\tabucline[0.5pt]{-}
ECO \cite{ECO17}&CVPR2017 &0.910&0.691&0.800&\textcolor{blue}{\textbf{0.605}}&0.741&0.537&\textcolor{blue}{\textbf{0.604}}&\textcolor{red}{\textbf{0.435}}&0.301&0.338&0.324&60&8\\\tabucline[0.5pt]{-}
%LMCF \cite{LMCF17}&CVPR2017 &-&0.568&-&-&-&-&-&-&-&-&-&61.38&8.11\\\tabucline[0.5pt]{-}
%ACFN \cite{ACFN17}&CVPR2017 &0.802&0.575&-&-&-&-&-&-&-&-&-&-&15\\\tabucline[0.5pt]{-}
%CREST \cite{CREST17}&ICCV2017 &0.837&0.623&0.730&0.554&-&-&-&-&-&-&-&-&1\\\tabucline[0.5pt]{-}
PTAV \cite{PTAV17}&ICCV2017 &0.849&0.635&0.741&0.544&-&-&\textcolor{red}{\textbf{0.624}}&\textcolor{blue}{\textbf{0.423}}&0.254&0.274&0.250&-&25\\\tabucline[0.5pt]{-}
DCFNET \cite{DCFNET17}&Arxiv2017 &0.816&0.601&-&-&-&-&-&-&-&-&-&-&\textcolor{blue}{\textbf{65.94}}\\\tabucline[0.5pt]{-}
%MCPF \cite{MCPF17}&CVPR2017 &0.873&0.628&0.774&0.545&-&-&-&-&-&-&-&-&1.80\\\tabucline[0.5pt]{-}
%MCPF+ \cite{MCPFs18}&PAMI2018 &0.887&0.643&0.779&0.571&-&-&-&-&-&-&-&-&0.5\\\tabucline[0.5pt]{-}
%HDT+ \cite{HDT+18}&PAMI2018 &0.912&0.687&-&-&-&-&-&-&-&-&-&-&1.4\\\tabucline[0.5pt]{-}
%MCCT \cite{MCCT18}&CVPR2018 &0.914&0.695&0.797&0.596&-&-&-&-&-&-&-&1.5&8.0\\\tabucline[0.5pt]{-}
STRCF \cite{STRCF18}&CVPR2018 &0.864&0.683&0.753&0.601&-&-&-&-&0.298&0.340&0.308&31.5&5.3\\\tabucline[0.5pt]{-}
UPDT \cite{UPDT18}&ECCV2018 &\textcolor{blue}{\textbf{0.931}}&\textcolor{red}{\textbf{0.702}}&-&\textcolor{red}{\textbf{0.622}}&-&0.550&-&-&-&-&-&-&-\\\tabucline[0.5pt]{-}
%LCT+ \cite{LCT+18}&IJCV2018 &0.825&0.592&-&-&-&-&-&-&-&-&-&13.8&-\\\tabucline[0.5pt]{-}
DSLT \cite{DSLT18}&ECCV2018 &0.909&0.660&\textcolor{blue}{\textbf{0.807}}&0.586&0.746&0.530&-&-&0.324&&0.323&-&5.7\\\tabucline[0.5pt]{-}
LSART \cite{LSART18}&CVPR2018 &0.923&0.672&-&-&-&-&-&-&-&-&-&-&1\\\tabucline[0.5pt]{-}
TRACA \cite{TRACA18}&CVPR2018 &0.816&0.603&-&-&-&-&-&-&0.227&0.278&0.257&-&\textcolor{red}{\textbf{101.3}}\\\tabucline[0.5pt]{-}
DRT \cite{DRT18}&CVPR2018 &0.923&\textcolor{blue}{\textbf{0.699}}&-&-&-&-&-&-&-&-&-&-&-\\\tabucline[0.5pt]{-}
HCFTs \cite{HCFTs19}&PAMI2019&0.870&0.598&-&-&-&-&-&-&-&-&-&-&6.70\\\tabucline[0.5pt]{-}
%ARCF \cite{ARCF19}&ICCV2019 &-&-&-&-&0.666&0.473&-&-&-&-&-&-&15.3\\\tabucline[0.5pt]{-}
GFS-DCF \cite{GFSDCF19}&ICCV2019 &\textcolor{red}{\textbf{0.932}}&0.693&-&-&-&-&-&-&&-&-&-&8\\\tabucline[0.5pt]{-}
RPCF \cite{RPCF19}&CVPR2019 &0.929&0.690&-&-&-&-&-&-&-&-&-&-&5\\\tabucline[0.5pt]{-}
ASRCF \cite{ASRCF19}&CVPR2019 &0.922&0.692&\textcolor{red}{\textbf{0.825}}&0.603&-&-&-&-&0.337&0.391&0.359&-&28\\\tabucline[0.5pt]{-}
ATOM \cite{ATOM19}&CVPR2019 &0.879&0.667&-&-&0.856&0.650&-&-&0.505&0.576&0.515&-&30\\\tabucline[0.5pt]{-}
DiMP \cite{DIMP19}&ICCV2019 &0.902&0.684&-&-&\textcolor{blue}{\textbf{0.858}}&\textcolor{blue}{\textbf{0.654}}&-&-&\textcolor{blue}{\textbf{0.563}}&\textcolor{red}{\textbf{0.650}}&\textcolor{blue}{\textbf{0.569}}&-&40\\\tabucline[0.5pt]{-}
%AutoTrack \cite{AUTO20}&CVPR2020 &-&-&-&-&0.671&0.477&-&-&-&-&-&-&59.2\\\tabucline[0.5pt]{-}
DSLT+ \cite{DSLT+20}&PAMI2020 &0.918&0.662&-&-&0.768&0.535&-&-&0.342&-&0.340&-&6.3\\\tabucline[0.5pt]{-}
PrDiMP \cite{PRDIMP20}&CVPR2020&0.903&0.696&-&-&\textcolor{red}{\textbf{0.878}}&\textcolor{red}{\textbf{0.680}}&-&-&\textcolor{red}{\textbf{0.609}}&-&\textcolor{red}{\textbf{0.598}}&-&30\\\tabucline[0.5pt]{-}
KYS \cite{KYS20}&ECCV2020 &-&0.695&-&-&-&-&-&-&-&\textcolor{blue}{\textbf{0.633}}&0.554&-&20\\\tabucline[2.5pt]{-}
\textbf{Siamese Trackers}&\textbf{Publication}&\textbf{PR}&\textbf{AUC}&\textbf{PR}&\textbf{AUC}&\textbf{PR}&\textbf{AUC}&\textbf{PR}&\textbf{AUC}&\textbf{PR}&\textbf{NPR}&\textbf{AUC}&\textbf{CPU}&\textbf{GPU}\\\tabucline[2.5pt]{-}
SiamFC \cite{SIAMFC16}&ECCV2016&0.771&0.582&0.688&0.503&0.691&0.461&0.613&0.399&0.339&0.420&0.336&-&86\\\tabucline[0.5pt]{-}
GOTURN \cite{GOTURN16}&ECCV2016&-&0.410&-&-&-&-&-&-&-&-&-&-&100\\\tabucline[0.5pt]{-}
SINT \cite{SINT16}&CVPR016&0.788&0.592&-&-&-&-&-&-&0.295&0.354&0.314&-&4\\\tabucline[0.5pt]{-}
CFNET \cite{CFNET17}&CVPR2017&0.748&0.568&0.607&0.441&0.651&0.436&0.570&0.349&0.259&0.312&0.275&-&75\\\tabucline[0.5pt]{-}
DSiam \cite{DSIAM17}&ICCV2017&-&-&0.711&0.505&-&-&-&-&0.322&0.405&0.333&-&45\\\tabucline[0.5pt]{-}
DaSiamRPN \cite{DASIAMRPN18}&ECCV2018&0.880&0.659&-&-&0.796&0.586&\textcolor{red}{\textbf{0.838}}&\textcolor{red}{\textbf{0.617}}&-&0.496&0.415&-&\textcolor{red}{\textbf{160}}\\\tabucline[0.5pt]{-}
RTINET \cite{RTINET18}&ECCV2018&-&0.682&-&\textcolor{blue}{\textbf{0.602}}&-&-&-&-&-&-&-&-&9\\\tabucline[0.5pt]{-}
SiamRPN \cite{SIAMRPN18}&CVPR2018&0.851&0.637&-&-&0.796&0.527&\textcolor{blue}{\textbf{0.617}}&\textcolor{blue}{\textbf{0.454}}&-&-&0.433&-&\textcolor{red}{\textbf{160}}\\\tabucline[0.5pt]{-}
GradNet \cite{GRADNET19}&ICCV2019&0.861&0.639&\textcolor{red}{\textbf{0.764}}&0.556&-&-&-&-&0.351&-&0.365&-&80\\\tabucline[0.5pt]{-}
SiamDW \cite{SIAMDW19}&CVPR2019&0.900&0.670&-&-&-&-&-&-&-&-&0.384&-&35\\\tabucline[0.5pt]{-}
TADT \cite{TADT19}&CVPR2019&0.866&0.660&-&0.562&-&-&-&-&-&-&-&-&33.7\\\tabucline[0.5pt]{-}
SiamRPN++ \cite{SIAMRPN19}&CVPR2019&0.914&0.696&-&-&0.807&0.613&-&-&0.491&0.569&0.496&-&35\\\tabucline[0.5pt]{-}
SPLT \cite{SPLT19}&ICCV2019&-&-&-&-&-&-&-&-&0.396&0.494&0.426&-&25.7\\\tabucline[0.5pt]{-}
SPM \cite{SPM19}&CVPR2019&0.899&0.687&-&-&-&-&-&-&-&-&-&-&120\\\tabucline[0.5pt]{-}
CLNET \cite{CLNET20}&ECCV2020&-&-&-&-&0.830&0.633&-&-&0.494&0.574&0.499&-&45.6\\\tabucline[0.5pt]{-}
Ocean \cite{OCEAN20}&ECCV2020&\textcolor{blue}{\textbf{0.920}}&0.684&-&-&-&-&-&-&\textcolor{blue}{\textbf{0.566}}&-&0.560&-&26\\\tabucline[0.5pt]{-}
ROAM \cite{ROAM20}&CVPR2020&0.904&0.680&-&-&-&-&-&-&0.445&-&0.447&-&20\\\tabucline[0.5pt]{-}
SiamBAN \cite{SIAMBAN20}&CVPR2020&0.910&0.696&-&-&0.833&0.631&-&-&0.521&0.598&0.514&-&40\\\tabucline[0.5pt]{-}
SiamFC++ \cite{SIAMFC20}&AAAI2020&-&0.683&-&-&-&-&-&-&-&-&0.544&-&60\\\tabucline[0.5pt]{-}
SiamR-CNN \cite{SIAMRCNN20}&CVPR2020&0.891&\textcolor{blue}{\textbf{0.701}}&-&\textcolor{red}{\textbf{0.649}}&\textcolor{blue}{\textbf{0.834}}&\textcolor{blue}{\textbf{0.649}}&-&-&\textcolor{red}{\textbf{0.684}}&\textcolor{red}{\textbf{0.722}}&\textcolor{red}{\textbf{0.648}}&-&4.7\\\tabucline[0.5pt]{-}
UDT++ \cite{UDT20}&IJCV2020&0.843&0.639&\textcolor{blue}{\textbf{0.725}}&0.552&-&-&-&-&-&-&0.305&-&55\\\tabucline[0.5pt]{-}
SiamAttn \cite{SIAMATTN20}&CVPR2020&\textcolor{red}{\textbf{0.926}}&\textcolor{red}{\textbf{0.712}}&-&-&\textcolor{red}{\textbf{0.845}}&\textcolor{red}{\textbf{0.650}}&-&-&-&\textcolor{blue}{\textbf{0.648}}&\textcolor{blue}{\textbf{0.560}}&-&45\\\tabucline[0.5pt]{-}
CSA \cite{CSA20}&CVPR2020&0.914&0.696&-&-&-&-&-&-&-&0.569&0.496&-&109\\\tabucline[0.5pt]{-}
D3S \cite{D3S20}&CVPR2020&0.841&0.608&-&-&-&-&-&-&0.490&-&0.488&-&25\\\tabucline[0.5pt]{-}
SiamCAR \cite{SIAMCAR20}&CVPR2020&0.910&0.698&-&-&0.760&0.614&-&-&0.510&0.600&0.507&-&52.27\\\tabucline[2.5pt]{-}
\end{tabu}
}}
\end{center}
\label{table1_dcf}
\vspace{-1.3\baselineskip}
\end{table*}

\begin{table}[t!]
\caption{Performance comparison of representative DCF and Siamese trackers, in terms of EAO, A, and R, on VOT2016 and VOT2018-ST. In case of DCF and Siamese trackers, the best two results are shown in red and blue fonts, respectively. }
\vspace{-1.8\baselineskip}
%\vspace{-1.0\baselineskip}
\begin{center}
\makebox[\linewidth]{
\scalebox{0.60}{
\begin{tabu}{|[2.0pt]c|[2.0pt]c|[2.0pt]c|c|c|[2.0pt]c|c|c|[2.0pt]c|[2.0pt]}
\tabucline[2.5pt]{3-16}
\multicolumn{2}{c|[2.0pt]}{}&\multicolumn{3}{c|[2.0pt]}{VOT2016}&\multicolumn{4}{c|[2.0pt]}{VOT2018-ST}\\\tabucline[2.5pt]{-}
\textbf{DCF Trackers}&\textbf{Publication}&\textbf{EAO$\uparrow$}&\textbf{A$\uparrow$}&\textbf{R$\downarrow$}&\textbf{EAO$\uparrow$}&\textbf{A$\uparrow$}&\textbf{R$\downarrow$}&\textbf{Speed}\\\tabucline[2.5pt]{-}
MOSSE \cite{MOSSE10}&CVPR2010&-&-&-&0.139&0.503&0.988&-\\\tabucline[0.5pt]{-}
%STC \cite{STC14}&ECCV2014&0.110&0.380&1.007&-&-&-&22.744\\\tabucline[0.5pt]{-}
SAMF \cite{SAMF14}&ECCV2014&0.186&0.507&0.587&0.093&0.484&1.302&4.099\\\tabucline[0.5pt]{-}
%ACA \cite{ACA14}&CVPR2014&0.173&0.446&0.662&-&-&-&18.26\\\tabucline[0.5pt]{-}
%DSST \cite{DSST14}&BMVC2014&0.181&0.533&0.704&0.079&0.395&1.452&12.747\\\tabucline[0.5pt]{-}
KCF \cite{KCF15}&PAMI2015&0.192&0.489&122&0.135&0.447&0.773&21.78\\\tabucline[0.5pt]{-}
HCF \cite{HCF15}&ICCV2015&0.220&0.450&0.396&-&-&-&1.057\\\tabucline[0.5pt]{-}
%SRDCF \cite{SRDCF15}&ICCV2015&0.248&0.535&0.419&0.119&0.490&0.974&1.99\\\tabucline[0.5pt]{-}
DeepSRDCF \cite{DEEPSRDCF15} &ICCV-W2015&0.276&0.528&0.326&0.154&0.492&0.707&0.38\\\tabucline[0.5pt]{-}
STAPLE \cite{STAPLE16}&CVPR2016&0.295&0.544&0.378&0.171&0.488&0.613&11.14\\\tabucline[0.5pt]{-}
%CCOT \cite{CCOT16}&ECCV2016&0.331&0.539&0.238&0.267&0.494&0.318&0.507\\\tabucline[0.5pt]{-}
%SCT \cite{SCT16}&CVPR2016&0.188&0.462&0.545&-&-&-&-\\\tabucline[0.5pt]{-}
%SRDCFdecon \cite{SRDCFDECON16}&CVPR2016&0.262&0.530&1.42&-&-&-&-\\\tabucline[0.5pt]{-}
%BACF \cite{BACF17}&ICCV2017&0.223&0.560&1.88&0.155&0.461&0.740&-\\\tabucline[0.5pt]{-}
%CSR-DCF \cite{CSRDCF17}&CVPR2017&0.338&0.524&0.239&0.256&0.491&0.356&-\\\tabucline[0.5pt]{-}
ECO \cite{ECO17}&CVPR2017&0.374&0.546&11.67&0.280&0.484&0.276&-\\\tabucline[0.5pt]{-}
%CREST \cite{CREST17}&ICCV2017&0.238&0.514&1.083&-&-&-&-\\\tabucline[0.5pt]{-}
DCFNET \cite{DCFNET17}&Arxiv2017&-&-&-&0.182&0.470&0.543&-\\\tabucline[0.5pt]{-}
%MCPF \cite{MCPF17}&CVPR2017&-&-&-&0.248&0.510&0.427&-\\\tabucline[0.5pt]{-}
%MCPF+ \cite{MCPFs18}&PAMI2018&-&-&-&0.257&0.513&0.397&-\\\tabucline[0.5pt]{-}
%HDT+ \cite{HDT+18}&PAMI2018&0.275&-&-&-&-&-&-\\\tabucline[0.5pt]{-}
%MCCT \cite{MCCT18}&CVPR2018&0.393&0.580&0.730&0.274&0.532&0.318&-\\\tabucline[0.5pt]{-}
STRCF \cite{STRCF18}&CVPR2018&0.313&0.550&0.920&0.345&0.523&0.215&-\\\tabucline[0.5pt]{-}
UPDT \cite{UPDT18}&ECCV2018&-&-&-&0.378&0.536&0.182&-\\\tabucline[0.5pt]{-}
DSLT \cite{DSLT18}&ECCV2018&0.332&0.541&15.48&0.325&0.543&0.224&-\\\tabucline[0.5pt]{-}
LSART \cite{LSART18}&CVPR2018&0.323&0.495&0.215&0.323&0.493&0.218&-\\\tabucline[0.5pt]{-}
TRACA \cite{TRACA18}&CVPR2018&-&-&-&0.137&0.424&0.857&-\\\tabucline[0.5pt]{-}
DRT \cite{DRT18}&CVPR2018&0.442&0.569&\textcolor{red}{\textbf{0.140}}&0.356&0.507&0.155&-\\\tabucline[0.5pt]{-}
GFS-DCF \cite{GFSDCF19}&ICCV2019&-&-&-&0.397&0.511&\textcolor{red}{\textbf{0.143}}&-\\\tabucline[0.5pt]{-}
RPCF \cite{RPCF19}&CVPR2019&-&-&-&0.316&0.500&0.234&-\\\tabucline[0.5pt]{-}
ASRCF \cite{ASRCF19}&CVPR2019&0.391&0.563&\textcolor{blue}{\textbf{0.187}}&0.328&0.494&0.234&-\\\tabucline[0.5pt]{-}
ATOM \cite{ATOM19}&CVPR2019&0.424&\textcolor{blue}{\textbf{0.617}}&0.190&0.401&0.590&0.204&-\\\tabucline[0.5pt]{-}
DiMP \cite{DIMP19}&ICCV2019&\textcolor{red}{\textbf{0.479}}&\textcolor{blue}{\textbf{0.617}}&0.190&0.440&0.597&\textcolor{blue}{\textbf{0.153}}&-\\\tabucline[0.5pt]{-}
DSLT+ \cite{DSLT+20}&PAMI2020&0.364&0.551&9.253&0.274&0.500&0.279&-\\\tabucline[0.5pt]{-}
PrDiMP \cite{PRDIMP20}&CVPR2020&\textcolor{blue}{\textbf{0.476}}&\textcolor{red}{\textbf{0.652}}&\textcolor{red}{\textbf{0.140}}&\textcolor{blue}{\textbf{0.442}}&\textcolor{red}{\textbf{0.618}}&0.165&-\\\tabucline[0.5pt]{-}
KYS \cite{KYS20}&ECCV2020&-&-&-&\textcolor{red}{\textbf{0.462}}&\textcolor{blue}{\textbf{0.609}}&\textcolor{red}{\textbf{0.143}}&-\\\tabucline[2.5pt]{-}
\textbf{Siamese Trackers}&\textbf{Publication}&\textbf{EAO$\uparrow$}&\textbf{A$\uparrow$}&\textbf{R$\downarrow$}&\textbf{EAO$\uparrow$}&\textbf{A$\uparrow$}&\textbf{R$\downarrow$}&\textbf{Speed}\\\tabucline[2.5pt]{-}
SiamFC \cite{SIAMFC16}&ECCV2016&0.235&0.532&0.461&0.188&0.503&0.585&9.213\\\tabucline[0.5pt]{-}
CFNET \cite{CFNET17}&CVPR2017&0.201&-&-&-&-&-&-\\\tabucline[0.5pt]{-}
DSiam \cite{DSIAM17}&ICCV2017&-&-&-&0.196&0.512&0.646&-\\\tabucline[0.5pt]{-}
DaSiamRPN \cite{DASIAMRPN18}&ECCV2018&0.411&0.610&0.22&0.326&0.570&0.337&-\\\tabucline[0.5pt]{-}
RTINET \cite{RTINET18}&ECCV2018&0.298&0.570&1.070&-&-&-&-\\\tabucline[0.5pt]{-}
SiamRPN \cite{SIAMRPN18}&CVPR2018&0.344&0.560&1.12&0.244&0.490&0.464&-\\\tabucline[0.5pt]{-}
GradNet \cite{GRADNET19}&ICCV2019&-&-&-&0.247&0.507&0.375&-\\\tabucline[0.5pt]{-}
SiamDW \cite{SIAMDW19}&CVPR2019&0.370&0.580&0.240&0.300&0.520&0.410&-\\\tabucline[0.5pt]{-}
SiamRPN++ \cite{SIAMRPN19}&CVPR2019&0.370&0.580&0.240&0.414&0.600&0.234&-\\\tabucline[0.5pt]{-}
TADT \cite{TADT19}&CVPR2019&0.299&0.550&1.170&-&-&-&-\\\tabucline[0.5pt]{-}
SPM \cite{SPM19}&CVPR2019&0.434&0.620&0.210&0.338&0.580&0.300&-\\\tabucline[0.5pt]{-}
Ocean \cite{OCEAN20}&ECCV2020&-&-&-&\textcolor{red}{\textbf{0.489}}&0.592&0.117&-\\\tabucline[0.5pt]{-}
ROAM \cite{ROAM20}&CVPR2020&0.441&0.599&0.174&0.380&0.543&0.195&-\\\tabucline[0.5pt]{-}
SiamBAN \cite{SIAMBAN20}&CVPR2020&\textcolor{blue}{\textbf{0.505}}&0.632&0.150&0.452&0.597&0.178&-\\\tabucline[0.5pt]{-}
SiamFC++ \cite{SIAMFC20}&AAAI2020&-&-&-&0.426&0.587&0.183&-\\\tabucline[0.5pt]{-}
SiamR-CNN \cite{SIAMRCNN20}&CVPR2020&0.461&0.645&0.173&0.408&0.609&0.220&-\\\tabucline[0.5pt]{-}
UDT++ \cite{UDT20}&IJCV2020&0.309&0.540&62&0.230&0.490&88&-\\\tabucline[0.5pt]{-}
SiamAttn \cite{SIAMATTN20}&CVPR2020&\textcolor{red}{\textbf{0.537}}&\textcolor{red}{\textbf{0.680}}&\textcolor{blue}{\textbf{0.140}}&\textcolor{blue}{\textbf{0.470}}&0.630&\textcolor{blue}{\textbf{0.160}}&-\\\tabucline[0.5pt]{-}
CSA \cite{CSA20}&CVPR2020&-&-&-&0.414&0.600&0.234&-\\\tabucline[0.5pt]{-}
D3S \cite{D3S20}&CVPR2020&0.493&\textcolor{blue}{\textbf{0.660}}&\textcolor{red}{\textbf{0.131}}&\textcolor{red}{\textbf{0.489}}&\textcolor{blue}{\textbf{0.640}}&\textcolor{red}{\textbf{0.150}}&-\\\tabucline[2.5pt]{-}
\end{tabu}
}}
\end{center}
\label{table2_dcf}
%\vspace{-1.3\baselineskip}
\end{table}

\begin{table}[t!]
\caption{Performance comparison of representative DCF and Siamese trackers, in terms of EAO, A, and R on VOT2020-ST. In case of DCF and Siamese trackers, the best two results are in red and blue fonts, respectively.}
\vspace{-1.3\baselineskip}
\begin{center}
\makebox[\linewidth]{
\scalebox{0.95}{
\begin{tabu}{|[2.0pt]c|[2.0pt]c|[2.0pt]c|c|c|[2.0pt]}
\tabucline[2.5pt]{3-13}
\multicolumn{2}{c|[2.0pt]}{}&\multicolumn{3}{c|[2.0pt]}{VOT2020-ST}\\\tabucline[2.5pt]{-}
\textbf{DCF Trackers}&\textbf{Publication}&\textbf{EAO$\uparrow$}&\textbf{A$\uparrow$}&\textbf{R$\uparrow$}\\\tabucline[2.5pt]{-}
KCF \cite{KCF15}&PAMI2015&0.154&0.407&0.432\\\tabucline[0.5pt]{-}
CSR-DCF \cite{CSRDCF17}&CVPR2017&0.193&0.406&0.582\\\tabucline[0.5pt]{-}
UPDT \cite{UPDT18}&ECCV2018&\textcolor{red}{\textbf{0.278}}&\textcolor{red}{\textbf{0.465}}&\textcolor{red}{\textbf{0.755}}\\\tabucline[0.5pt]{-}
DiMP \cite{DIMP19}&ICCV2019&\textcolor{blue}{\textbf{0.274}}&0.457&\textcolor{blue}{\textbf{0.740}}\\\tabucline[0.5pt]{-}
ATOM \cite{ATOM19}&CVPR2019&0.271&\textcolor{blue}{\textbf{0.462}}&0.734\\\tabucline[2.5pt]{-}
\textbf{Siamese Trackers}&\textbf{Publication}&\textbf{EAO$\uparrow$}&\textbf{A$\uparrow$}&\textbf{R$\uparrow$}\\\tabucline[2.5pt]{-}
SiamFC \cite{SIAMFC16}&ECCV2016&0.179&0.418&0.502\\\tabucline[0.5pt]{-}
SiamMask \cite{SIAMMASK19}&CVPR2019&0.321&0.624&0.648\\\tabucline[0.5pt]{-}
D3S \cite{D3S20}&CVPR2020&\textcolor{red}{\textbf{0.439}}&\textcolor{red}{\textbf{0.699}}&\textcolor{red}{\textbf{0.769}}\\\tabucline[0.5pt]{-}
Ocean \cite{OCEAN20}&ECCV2020&\textcolor{blue}{\textbf{0.430}}&\textcolor{blue}{\textbf{0.693}}&\textcolor{blue}{\textbf{0.754}}\\\tabucline[2.5pt]{-}
\end{tabu}
}}
\end{center}
\label{table3_seg}
\end{table}

\begin{table}[t!]
\caption{Performance comparison of representative DCF and Siamese trackers, in terms of mean Average Overlape (mAO), mSR$_{0.50}$, and mSR$_{0.75}$, PR, NPR, and AUC, on GOT-10K and TrackingNet. In case of DCF and Siamesetrackers, the best two results are in red and blue fonts, respectively.}
%\vspace{-1.0\baselineskip}
\vspace{-1.3\baselineskip}
\begin{center}
\makebox[\linewidth]{
\scalebox{0.60}{
\begin{tabu}{|[2.0pt]c|[2.0pt]c|[2.0pt]c|c|c|[2.0pt]c|c|c|[2.0pt]}
\tabucline[2.5pt]{3-16}
\multicolumn{2}{c|[2.0pt]}{}&\multicolumn{3}{c|[2.0pt]}{GOT-10K}&\multicolumn{3}{c|[2.0pt]}{TrackingNet}\\\tabucline[2.5pt]{-}
\textbf{DCF Trackers}&\textbf{Publication}&\textbf{mAO}&\textbf{mSR$_{0.50}$}&\textbf{mSR$_{0.75}$}&\textbf{PR}&\textbf{NPR}&\textbf{AUC}\\\tabucline[2.5pt]{-}
MOSSE \cite{MOSSE10}&CVPR2010&-&-&-&0.326&0.442&0.388\\\tabucline[0.5pt]{-}
%%CSK \cite{CSK12}&ECCV2012&0.264&0.236&0.085&0.368&0.503&0.429\\\tabucline[0.5pt]{-}
SAMF \cite{SAMF14}&ECCV2014&0.330&0.331&0.130&0.477&0.598&0.504\\\tabucline[0.5pt]{-}
%%DSST \cite{DSST14}&BMVC2014&0.317&0.314&0.136&0.460&0.588&0.464\\\tabucline[0.5pt]{-}
KCF \cite{KCF15}&PAMI2015&0.279&0.263&0.099&0.419&0.546&0.447\\\tabucline[0.5pt]{-}
HCF \cite{HCF15}&ICCV2015&0.379&0.380&0.134&-&-&-\\\tabucline[0.5pt]{-}
%SRDCF \cite{SRDCF15}&ICCV2015&0.312&0.310&0.134&0.455&0.573&0.521\\\tabucline[0.5pt]{-}
STAPLE \cite{STAPLE16}&CVPR2016&0.332&0.333&0.135&0.470&0.603&0.528\\\tabucline[0.5pt]{-}
%CCOT \cite{CCOT16}&ECCV2016&0.406&0.415&0.161&-&-&-\\\tabucline[0.5pt]{-}
fDSST \cite{fDSST16}&PAMI2016&0.289&0.278&0.121&-&-&-\\\tabucline[0.5pt]{-}
%%SRDCFdecon \cite{SRDCFDECON16}&CVPR2016&0.310&0.311&0.139&-&-&-\\\tabucline[0.5pt]{-}
%BACF \cite{BACF17}&ICCV2017&0.346&0.361&0.149&0.461&0.580&0.523\\\tabucline[0.5pt]{-}
%CSR-DCF \cite{CSRDCF17}&CVPR2017&-&-&-&0.480&0.622&0.534\\\tabucline[0.5pt]{-}
ECO \cite{ECO17}&CVPR2017&0.395&0.407&0.170&0.492&0.618&0.554\\\tabucline[0.5pt]{-}
%%CACF \cite{CACF17}&CVPR2017&-&-&-&0.468&0.605&0.529\\\tabucline[0.5pt]{-}
DCFNET \cite{DCFNET17}&Arxiv2017&0.364&0.378&0.144&-&-&-\\\tabucline[0.5pt]{-}
STRCF \cite{STRCF18}&CVPR2018&0.449&0.481&0.169&-&-&-\\\tabucline[0.5pt]{-}
UPDT \cite{UPDT18}&ECCV2018&-&-&-&0.557&0.703&0.611\\\tabucline[0.5pt]{-}
GFS-DCF \cite{GFSDCF19}&ICCV2019&-&-&-&0.565&0.717&0.609\\\tabucline[0.5pt]{-}
ATOM \cite{ATOM19}&CVPR2019&0.556&0.634&0.402&\textcolor{blue}{\textbf{0.648}}&\textcolor{blue}{\textbf{0.771}}&\textcolor{blue}{\textbf{0.703}}\\\tabucline[0.5pt]{-}
DiMP \cite{DIMP19}&ICCV2019&\textcolor{blue}{\textbf{0.611}}&\textcolor{blue}{\textbf{0.717}}&\textcolor{blue}{\textbf{0.492}}&\textcolor{red}{\textbf{0.800}}&\textcolor{red}{\textbf{0.801}}&\textcolor{red}{\textbf{0.740}}\\\tabucline[0.5pt]{-}
PrDiMP \cite{PRDIMP20}&CVPR2020&\textcolor{red}{\textbf{0.634}}&\textcolor{red}{\textbf{0.738}}&\textcolor{red}{\textbf{0.543}}&0.704&0.816&0.758\\\tabucline[0.5pt]{-}
KYS \cite{KYS20}&ECCV2020&-&-&-&&0.633&0.554\\\tabucline[2.5pt]{-}
\textbf{Siamese Trackers}&\textbf{Publication}&\textbf{mAO}&\textbf{mSR$_{0.50}$}&\textbf{mSR$_{0.75}$}&\textbf{PR}&\textbf{NPR}&\textbf{AUC}\\\tabucline[2.5pt]{-}
SiamFC \cite{SIAMFC16}&ECCV2016&0.392&0.426&0.135&0.533&0.654&0.571\\\tabucline[0.5pt]{-}
GOTURN \cite{GOTURN16}&ECCV2016&0.418&0.475&0.163&-&-&-\\\tabucline[0.5pt]{-}
CFNET \cite{CFNET17}&CVPR2017&0.434&0.481&0.190&0.533&0.654&0.578\\\tabucline[0.5pt]{-}
DSiam \cite{DSIAM17}&ICCV2017&0.417&0.461&0.149&-&0.405&0.333\\\tabucline[0.5pt]{-}
DaSiamRPN \cite{DASIAMRPN18}&ECCV2018&0.444&0.536&0.220&0.591&0.733&0.638\\\tabucline[0.5pt]{-}
SiamRPN \cite{SIAMRPN18}&CVPR2018&0.481&0.581&0.270&-&-&-\\\tabucline[0.5pt]{-}
SiamDW \cite{SIAMDW19}&CVPR2019&0.411&0.456&0.154&-&-&-\\\tabucline[0.5pt]{-}
SiamRPN++ \cite{SIAMRPN19}&CVPR2019&0.518&0.618&0.325&0.694&0.800&0.733\\\tabucline[0.5pt]{-}
SPM \cite{SPM19}&CVPR2019&0.513&-&-&0.660&0.771&0.712\\\tabucline[0.5pt]{-}
Ocean \cite{OCEAN20}&ECCV2020&\textcolor{blue}{\textbf{0.611}}&\textcolor{blue}{\textbf{0.721}}&-&-&-&-\\\tabucline[0.5pt]{-}
ROAM \cite{ROAM20}&CVPR2020&0.465&0.532&0.236&0.623&0.754&0.670\\\tabucline[0.5pt]{-}
SiamFC++ \cite{SIAMFC20}&AAAI2020&0.595&0.695&\textcolor{blue}{\textbf{0.479}}&0.694&0.800&0.733\\\tabucline[0.5pt]{-}
SiamR-CNN \cite{SIAMRCNN20}&CVPR2020&\textcolor{red}{\textbf{0.649}}&\textcolor{red}{\textbf{0.728}}&\textcolor{red}{\textbf{0.597}}&\textcolor{red}{\textbf{0.800}}&\textcolor{red}{\textbf{0.854}}&\textcolor{red}{\textbf{0.812}}\\\tabucline[0.5pt]{-}
UDT++ \cite{UDT20}&IJCV2020&-&-&-&0.495&0.633&0.563\\\tabucline[0.5pt]{-}
SiamAttn \cite{SIAMATTN20}&CVPR2020&-&-&-&\textcolor{blue}{\textbf{0.715}}&\textcolor{blue}{\textbf{0.817}}&\textcolor{blue}{\textbf{0.752}}\\\tabucline[0.5pt]{-}
D3S \cite{D3S20}&CVPR2020&0.597&0.676&0.462&0.664&0.768&0.728\\\tabucline[0.5pt]{-}
SiamCAR \cite{SIAMCAR20}&CVPR2020&0.569&0.670&0.415&-&-&-\\\tabucline[2.5pt]{-}
\end{tabu}
}}
\end{center}
\label{table3_dcf}
%\vspace{-1.3\baselineskip}
\end{table}

\subsection{Speed Comparison}
The tracking speed is another very important metric to evaluate the trackers especially to meet the real-time requirements. 
However, evaluating the tracking speed is not straightforward since a number of key-factors influence including feature extraction, model update method, programming language, and most importantly the hardware that the trackers are implemented on.
To reduce the influence of the hardware, the VOT2014 committee has introduced a new unit known as Equivalent Filter Operations (EFO) that reports the tracking speed in terms of a predefined filtering operations that the toolkit automatically carries out prior to running the experiments \cite{VOT2014}.
Tables \ref{table2_dcf} presents the tracking speed in terms of EFO measures of SOTA trackers.
Trackers KCF and STAPLE clearly show the best speed while deep trackers including DeepSRDCF and HCF show competitive performance but worst speed.

\section{Discussion and Conclusions}
\label{sec:maindiscussion}
Here, we summarize the common lessons learned and a set of recommendations for future work in generic object tracking. 

\newcommand{\parsection}[1]{\noindent\textbf{#1}}

\parsection{Importance of end-to-end tracking framework:} These frameworks have demonstrated excellent performance recently. While end-to-end offline learning is a prerequisite for Siamese tracking, recent DCF methods have also adopted this paradigm with success. Thus, learning the underlying features, along with prediction heads, have proven crucial for optimal performance. This has only been possible in the last few years, with the introduction of large scale training datasets.

\parsection{Importance of robust target modeling:}
While Siamese-based approaches have excelled in many areas, end-to-end DCF-based methods still demonstrate an advantage in challenging long-term tracking scenarios, such as LaSOT. This shows the importance of robust online target appearance modelling, achieved by embedding discriminative learning modules into the network architecture. Such methods effectively integrate background appearance cues and can easily be updated during the tracking procedure using online learning.

\parsection{Target state estimation:}
Siamese-based approaches have driven the advancement of more accurate bounding box regression by leveraging progress in the neighboring field of object detection. Recent one-stage (anchor free) based approaches, such as Ocean, achieve simple, accurate, and efficient bounding box regression. Furthermore, these strategies are generic and can easily be integrated in any visual tracking architecture. 

\parsection{Role of segmentation:}
Although the task of bounding box regression have seen substantial progress in tracking, such a target state model is inherently limited. Instead, segmentation promises a pixel-precise estimation of the target, which is highly desired in many applications. Moreover, segmentation offers the potential of improving the tracking itself, by for example aiding the target model update. Furthermore, as demonstrated in for instance \cite{ALPHA21}, segmentation further aids the regression of accurate bounding boxes and helps estimating the scale at which the object is tracked. Future efforts should therefore be aimed towards integrating accurate segmentation into robust tracking frameworks.

\parsection{Backbone architectures:}
The ResNet architecture has withstood the test of time in several computer vision applications. In visual tracking, it remains the most popular choice for feature extraction. The architecture is simple, effective, and allows for the extraction of features at multiple resolutions. While effective when pushing the boundaries of SOTA, it is still remains computationally costly for real-time applications on platforms with harder computational constraints, such as CPUs. A highly interesting future direction is therefore to develop efficient backbone networks tailored for the tracking task \cite{yan2021lighttrack}.

\parsection{Estimating geometry:} In some applications, e.g.\ in augmented reality, a precise geometric transformation between frames is necessary for the added graphics to appear as attached to objects. For planar objects, at least an affine transformation, but preferably a homography, between a reference and current view is required. For non-planar objects, the problems is linked to online reconstruction of the 3D shape of the object.  Neither DCF nor Siamese methods have been equipped to provide precise geometric correspondence, which remains an open research issue.

\parsection{Role of Transformers:}
Transformers have recently shown success in a variety of vision tasks \cite{dosovitskiy2020image}. 
Very recent tracking approaches employ transformers in different ways. \cite{chen2021transformer, wang2021transformer} utilize transformers for feature enhancement, in combination with either DCF of Siamese trackers.
\cite{mayer2021learning} employs a transformer to associate the target object between frames in the presence of distractors.
In particular, STARK employs a transformer module for target detection and bounding box regression \cite{yan2021learning}. 
In this work, the transformer thus takes on the role of the DCF or Siamese correlation component. 
The transformer, with its embedded attention module, bares interesting commonalities with DCF. Most importantly, it allows for integration of background appearance information through global operations. 
Moreover, the transformer employed in STARK predicts a correlation filter. It therefore can be seen as a replacement of the optimization based filter prediction in DCF. Much future effort is needed to further analyze the effectiveness of transformers, as well as their relation to the DCF and Siamese paradigms.

\parsection{Future directions:} In the recent past, with the introduction of precise segmentation capabilities to visual trackers, the connection to video segmentation has become apparent. 
Some top visual tracking methods perform well on video segmentation benchmarks \cite{D3S20, SIAMMASK19}, despite the restriction to causal processing of the input. 

In the near future, we expect convergence with areas as SLAM, which estimates the relative position of the camera model of the scene that is build online, assuming a rigid scene. As soon as the rigidity assumption is dropped, as in multi-body SLAM, the output for the object not corresponding to the background can be seen as tracking of the object, with an estimated 3D shape model added.

The steadily improving performance of single-object tracking, including robustness to nearly identical distractors, demonstrated e.g., in sequences of groups of insects, seem like to lead to convergence with the research in the area of multi-target tracking. We are probably likely to see strong progress in the Open-world Tracking Problem, as described in \cite{liu2021opening}, i. e., of methods capable of tracking, segmenting, and shape modelling of multiple objects from a-priori unknown classes.

%\section*{Acknowledgments} 
%This publication acknowledges the support provided by the Khalifa University of Science and Technology under Award No. RC1-2018-KUCARS.

\ifCLASSOPTIONcaptionsoff
  \newpage
\fi

\bibliographystyle{IEEEtranS}
\bibliography{bare_jrnl}

\end{document}